\pdfoutput=1

\documentclass[11pt]{article}

\usepackage[final]{acl}

\usepackage{times}
\usepackage{latexsym}

\usepackage[T1]{fontenc}

\usepackage[utf8]{inputenc}

\usepackage{microtype}

\usepackage{inconsolata}

\usepackage{graphicx}

\title{What's the Difference? Supporting Users in Identifying the Effects of Prompt and Model Changes Through Token Patterns}

\author{
  \textbf{Michael A. Hedderich\textsuperscript{1}},
  \textbf{Anyi Wang\textsuperscript{1}}, %
  \textbf{Raoyuan Zhao\textsuperscript{1}},\\
  \textbf{Florian Eichin\textsuperscript{1}},
   \textbf{Jonas Fischer\textsuperscript{2}},
  \textbf{Barbara Plank\textsuperscript{1}}
\\
\textsuperscript{1}MaiNLP, Center for Information and Language Processing, LMU Munich \\
Munich Center for Machine Learning (MCML)
\\
\textsuperscript{2}Department for Computer Vision and Machine Learning, Max Planck Institute for Informatics \\
\texttt{\{mhedderich,\,raoyuan.zhao,\,feichin,\,b.plank\}@cis.lmu.de},\\ \texttt{anyi.wang@campus.lmu.de,jonas.fischer@mpi-inf.mpg.de}
}

\usepackage{xspace}
\usepackage{amsmath}

\usepackage{pifont}
\usepackage{subfig}

\newcommand{\ourmethod}{\texttt{Spotlight}\xspace}
\newcommand{\genoutput}{\ensuremath{g}\xspace}
\newcommand{\gensourceone}{\ensuremath{G_1}\xspace}
\newcommand{\gensourcetwo}{\ensuremath{G_2}\xspace}
\newcommand{\gensourceonetwo}{\ensuremath{G_{1/2}}\xspace}
\newcommand{\gensourceorig}{\ensuremath{G_o}\xspace}
\newcommand{\gensourceorigone}{\ensuremath{G_o^1}\xspace}
\newcommand{\gensourceorigtwo}{\ensuremath{G_o^2}\xspace}
\newcommand{\transformationfunction}{\ensuremath{t}\xspace}

\newcommand{\pattern}{\ensuremath{p}\xspace}

\newcommand{\tp}[1]{$\{$\textit{#1}$\}$}
\newcommand{\prompt}[1]{\texttt{#1}}

\definecolor{DarkGreen}{RGB}{1,180,32}
\newcommand{\cmark}{{\color{DarkGreen}\ding{51}}}
\newcommand{\xmark}{{\color{red}\ding{55}}}
\newcommand{\lmark}{{\color{gray}-}}

\makeatletter
\newcommand\incircbin
{%
  \mathpalette\@incircbin
}
\newcommand\@incircbin[2]
{%
  \mathbin%
  {%
    \ooalign{\hidewidth$#1#2$\hidewidth\crcr$#1\bigcirc$}%
  }%
}

\begin{document}
\maketitle
\begin{abstract}

Prompt engineering for large language models is challenging, as even small prompt perturbations or model changes can significantly impact the generated output texts. Existing evaluation methods of LLM outputs, either automated metrics or human evaluation, have limitations, such as providing limited insights or being labor-intensive. We propose \ourmethod, a new approach that combines both automation and human analysis. Based on data mining techniques, we automatically distinguish between random (decoding) variations and \emph{systematic} differences in language model outputs. This process provides token patterns that describe the systematic differences and guide the user in manually analyzing the effects of their prompts and changes in models efficiently. We create three benchmarks to quantitatively test the reliability of token pattern extraction methods and demonstrate that our approach provides new insights into established prompt data. From a human-centric perspective, through demonstration studies and a user study, we show that our token pattern approach helps users understand the systematic differences of language model outputs. We are further able to discover relevant differences caused by prompt and model changes (e.g.\ related to gender or culture), thus supporting the prompt engineering process and human-centric model behavior research.

\end{abstract}

\section{Introduction}

Prompting has become a key method for controlling large language models (LLMs)~\cite{Liu23Prompting}, but finding the right prompt (prompt engineering) is still challenging. While prompting allows users to influence various aspects of the generated output covering both content and style, even a semantically insignificant change in the prompt can have a substantial effect on the generated output~\cite{mizrahi-etal-2024-state,shu-etal-2024-dont}. This makes finding the right prompt for a task difficult, especially in complex scenarios and when carried out by non-AI-experts~\cite{Zamfirescu-Pereira2023HerdingAICats,Zamfirescu-Pereira2023Johnny}.

Different approaches exist to evaluate the output generated by LLMs and to understand the effects of prompt and model changes. Automated metrics, such as accuracy or BLEU scores~\cite{papineni-etal-2002-bleu}, are fast to compute and allow for easy quantitative comparisons. However, such task-specific metrics only measure predefined criteria and just provide a simplified view on the long texts that modern LLMs generate. In contrast, human annotators can analyze texts in a nuanced and flexible way, but this process is labor-intensive~\cite{Hopkins2021OutsideBigTech} and prone to errors~\cite{wu2023stylesubstanceevaluationbiases,ghosh2024comparedespairreliablepreference}, especially in light of large amounts of LLM outputs. A core challenge when analyzing LLM outputs is their stochastic decoding~\cite{Holtzman2020NucleusSampling}. Even for the same model and prompt, two LLM outputs are likely different from each other. To distinguish systematic differences from random decoding variations, analyzing large amounts of LLM outputs is hence required when comparing two prompts or models.

\begin{figure*}
    \centering
    \includegraphics[width=0.98\textwidth]{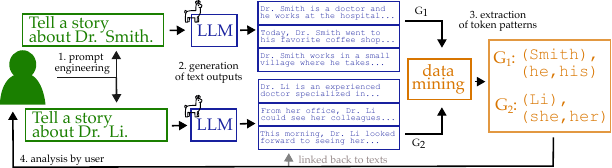}
    \caption{Workflow of our proposed \ourmethod approach when users compare the effects of a prompt change, which aids the detection of e.g.\ gender bias. Alternatively, different models using the same prompt can be analyzed.}
    \label{fig:figure1}
\end{figure*}

In this work, we propose to combine the efficiency and quantification of automated methods with the flexibility of human evaluators, as visualized in Figure~\ref{fig:figure1}. In a nutshell, our approach works as follows: We generate a set of outputs for two different prompts or models and then automatically extract descriptions of the differences between the sets of texts. By adapting data mining techniques for this task, we leverage their reliability in identifying systematic differences. The differences are described as sets of tokens (token patterns) that offer a concise form while preserving nuances. The token patterns then guide the analysis of a human evaluator and help them understand the effects of their prompt and model changes. We call our approach \ourmethod, as it shines light on Systematic Pattern differences in Output Tokens.

We see \ourmethod not as a replacement for existing, established evaluation techniques such as automatic metrics and full manual evaluation. Rather, we propose a middle ground and want to provide a complementary way to support users in prompt engineering and researchers in model evaluation. In contrast to existing behavioral testing like CheckList~\cite{ribeiro-etal-2020-beyond}, we are interested in generated, longer LLM output texts and do not depend on a single target metric. 

Through a set of diverse experiments covering both quantitative benchmarks and human-centric evaluations, including a user study, we show that \ourmethod  provides human-understandable and faithful explanations which help users better understand the differences between groups of generated texts within and across LLMs, and that it is able to uncover unknown effects of prompt and model changes, relating, e.g., to gender biases, cultural artifacts and task failure.

Our contributions are as follows:
\begin{itemize}
    \item We propose \ourmethod, a token-pattern-based approach to identify  systematic differences caused by prompt and model changes.
    \item We create three quantitative benchmarks with known ground truth to evaluate how well counting-, data mining- and LLM-based techniques can identify systematic differences in groups of LLM outputs.
    \item We show that our approach discovers more fine-grained differences on established prompt data than previous work.
    \item We demonstrate the ability to discover unexpected LLM behavior, uncovering, e.g., for story generation, how changing the last name of a doctor drastically impacts the gender distribution in the text.
    \item We show in a user study the helpfulness of the token patterns to human evaluators.
\end{itemize}

The code and data for this work, including the benchmarks and all token pattern results, are publicly available at \url{https://github.com/mainlp/spotlight}.

\section{Related Work}

\subsection{Prompt \& Model Sensitivity}

In tuning-free prompting, a user provides instructions and possibly examples (in-context learning) to achieve their desired LLM output~\cite{Liu23Prompting}. How the prompt is formulated can have a large and unexpected impact on the generated outputs. \citet{gu-etal-2023-robustness} showed that performance can deteriorate when prompts are changed, e.g., when deleting stopwords. \citet{mizrahi-etal-2024-state} analyzed 20 LLMs and 39 tasks, finding that results are unstable when prompts are paraphrased. Similarly, for prompts asking for opinions, even semantically insignificant changes like adding whitespaces can impact the outputs~\cite{shu-etal-2024-dont}. \citet{webson-pavlick-2022-prompt} found evidence that, to a certain degree, LLMs understand task instructions differently than humans. Actual users, especially non-AI-experts, often struggle with prompting, e.g., because they overgeneralize from single examples~\cite{Zamfirescu-Pereira2023Johnny} or because a prompt change might trigger unintended outputs~\cite{Zamfirescu-Pereira2023HerdingAICats}.

\subsection{Metrics for Text Evaluation}

Task-specific metrics for comparing model outputs can be calculated automatically, making them cost-effective and easy to apply, but they often have limitations. Metrics such as accuracy and F1 scores require a fixed label format and can thus not be directly applied to longer and more variable LLM-generated texts. Furthermore, they only measure predefined criteria and may overlook unexpected differences. For automated text comparison metrics such as BLEU~\cite{papineni-etal-2002-bleu} or ROUGE~\cite{lin-2004-rouge} there is substantial work that highlights their flaws, like low correlation with human judgments~\cite{Rankel2013AutomaticContentEvaluation, reiter2018reviewBleu,tay-etal-2019-red,fabbri2020summeval,AlvaManchego2021UnsuitabilityAutomaticEvaluation,goyal2023newssummarizationevaluationera}. The practitioners in \cite{gero2024supportingsensemakinglargelanguage} reported that they found the automatic evaluations not useful. An additional challenge for label and text metrics is the need to create evaluation or reference datasets, which our approach does not suffer from. 

\subsection{Human Evaluation}

Human annotators can manually analyze and compare outputs, a process often considered the gold standard, but this is labor-intensive and error-prone. Experts can evaluate both the main task goals and unforeseen issues, but this might require a complex evaluation process~\cite{Cabrera2023DataScientistsMakeSense} and might be limited by resource constraints~\cite{Hopkins2021OutsideBigTech}. Outsourcing the analysis to crowdworkers comes with its own challenges, such as maintaining quality~\cite{Hettiachchi2022TaskAssignmentCrowdsourcing}. In general, human evaluators might be inconsistent~\cite{ghosh2024comparedespairreliablepreference}, be influenced by order bias~\cite{wu2023stylesubstanceevaluationbiases}, or find patterns in data where none exist~\cite{ellerby2017apophenia}. Two goals of our approach are to reduce manual effort and limit false manual conclusions by only providing token patterns to the user that indicate systematic differences.

\subsection{LLM as Judge}
Recently, the use of LLMs to evaluate LLM-generated outputs has gained popularity due to them being automatic and often reference-free~\cite{wu-etal-2023-holistic,akkasi-etal-2023-reference,gao2024llmbasednlgevaluationcurrent,chang2024SurveyEvaluation,li-etal-2024-leveraging-large, xu-etal-2023-instructscore}. These approaches also suffer, however, from biases, instability, and lack of correlation with human judgments~\cite{bavaresco2024llmsinsteadhumanjudges,wang-etal-2024-large-language-models-fair,wu2023stylesubstanceevaluationbiases,deutsch-etal-2022-limitations,li-etal-2024-leveraging-large}. The risk of hallucinations~\cite{huang2025Hallucinations} and their low performance in math and reasoning tasks~\cite{cobbe2021trainingverifierssolvemath,srivatsa-kochmar-2024-makes} further exacerbate issues in our setting where we search for systematic differences between texts. We evaluate the suitability of LLMs as judges in our benchmarks.

\subsection{Interdisciplinary Evaluations of NLP Models}

To evaluate a model in more depth than just with a single accuracy metric, the CheckList approach~\cite{ribeiro-etal-2020-beyond} evaluates failure rates for a fixed set of linguistic capabilities (behavioral testing). SynthEval~\cite{zhao-etal-2024-syntheval} creates such test sets automatically using multiple NLP models. Dynabench~\cite{kiela-etal-2021-dynabench} asks users to come up with difficult test cases, and the work by \citet{ribeiro-lundberg-2022-adaptive} supports the user with LLM suggestions for new test cases. These works share our idea of identifying conceptual structures that can be analyzed by users. They focus, however, on testing single models that make label predictions, while we are interested in comparing between models that generate longer texts and we identify differences that are not dependent on a single target metric.

\citet{cheng-etal-2023-marked} identify words that occur differently in the outputs for prompts that contain specific gender and ethnicity words. However, their approach is limited to the identification of single words.

Understanding the large output space of models can be seen as a sensemaking problem~\cite{pirolli2005sensemaking,gero2022sensemaking}, a concept from the field of human-computer interaction. User interfaces can support this process, e.g.\ by grouping outputs according to predefined rules~\cite{gero2022sensemaking} or by highlighting significant words or sequences within each individual output text~\cite{gero2024supportingsensemakinglargelanguage}. In these works, the user is still supposed to inspect a large set of outputs while we provide concise support to the user through succinct token patterns. Zeno~\cite{cabrera23Zeno} and ChainForge~\cite{Arawjo24ChainForge} are platforms for visualizing and testing AI models and LLMs. Our approach could be naturally integrated into such platforms.

\section{Mining Systematic LLM Output Differences}

Our goal is to identify systematic differences between generated texts from two different sources, e.g., from two different prompts for the same LLM or the same prompt provided to two different LLMs.  Identifying these differences is non-trivial as even two generated output texts using the \textit{same source} (the same prompt and model) will likely not be identical. This is due to stochastic effects in the decoding methods which are essential for high-quality and human-like texts~\cite{Holtzman2020NucleusSampling}. We, therefore, need to ignore the differences \textit{between the sources} that are due to \textit{random decoding variations} and only identify those differences that are \textit{systematic and consistent} across many samples of the sources.

Identifying these differences through manual reading of large amounts of textual outputs does not scale. We want to reduce the manual effort while at the same time not reducing the analysis to a simplistic metric that might miss crucial aspects. Specifically, we are interested in a succinct and faithful explanation of the systematic differences that can be directly linked back to and further inspected in the original data. For this, we choose sets of tokens (token patterns) as description medium. For example, a gender bias could thus be indicated by the token set \tp{he, him} occurring much more frequently in one of the two sources. Similarly, a change in formality might be identified by token pattern phrases such as \tp{Dear, Professor}.

Formally, let \genoutput be a token sequence. Let \gensourceone and \gensourcetwo be two sets of such generated sequences obtained from different sources. A token pattern \pattern is a set of tokens and we say that \pattern occurs in \genoutput if $\pattern \subseteq set(\genoutput)$. A token pattern might occur in one or multiple token sequences in \gensourceone and \gensourcetwo. We are interested in those token patterns whose number of occurrences is systematically different between the sequences in \gensourceone and \gensourcetwo.

\section{Token Pattern Discovery Methods}
\label{sec:token-discovery-methods}

\ourmethod is independent of a specific method to identify the token patterns and can be instantiated using different approaches. We compare counting-, data mining- and LLM-based techniques on our benchmarks. For counting and identifying systematically different token distributions, tf-idf~\cite{sparck1972statistical} is a very established approach in NLP. We evaluate \textbf{c-tf-idf} by \citet{grootendorst2022bertopic} which is an adaptation that distinguishes between classes (here, two groups of texts). 

Data mining methods extend the counting idea with additional techniques for more robustness (e.g.\ by adding significance testing) and for more complex patterns (e.g.\ by adding search algorithms). We adapt four different data mining techniques to our setting. \textbf{Premise}~\cite{hedderichfischer2022LabelDescriptivePatterns} uses lossless compression to find non-redundant token patterns that distinguish two groups of text. \textbf{SPuManTE}~\cite{pellegrina19Spumante} selects patterns based on statistical testing. \textbf{Cortana}~\cite{Meeng2011Cortana} and \textbf{PySubgroup}~\cite{lemmerich2018pysubgroup} both belong to the subgroup discovery methods which identify subgroups (in our case token patterns) %
that differ in distribution. 

To compare to LLM-based judging, we evaluate two versions of the open source model \textbf{Llama 3.1} (8B and 70B) \cite{grattafiori2024llama3herdmodels} as well as the commercial \textbf{GPT-4o} ~\cite{openai2024gpt4ocard}.

We use our benchmarks to select the best method to identify the token patterns. We use that best method for the in-depth evaluations of our general \ourmethod approach afterward. Some methods do not scale to our need to process large groups of generated LLM outputs, either due to memory or maximum token length restrictions, or due to extremely long runtimes. We exclude these methods from some benchmarks with details below. More information about the methods and our adaptations to the token pattern task are in Appendix~\ref{sec:appendix_methods}. 

\section{Experiments with Known Ground Truth}
\label{sec:exp_ground_truth}
In real use cases, the ground truth is usually not known due to the blackbox nature of LLMs, making proper evaluation difficult. Therefore, we create three benchmark datasets where the differences between the sets of texts (\gensourceone and \gensourcetwo) are injected synthetically into the texts, thus introducing an exact ground truth. We make sure to obtain realistic textual settings by taking inspiration from existing NLP work.

\subsection{Benchmarks}

For each dataset, we generate a set of text sequences \gensourceorig through an LLM using the same prompt but different randomization in the decoding, thus obtaining different sequences from the same text distribution. This set of sequences is divided into two halves \gensourceorigone and \gensourceorigtwo. We define two transformation functions $\transformationfunction_1,\transformationfunction_2$ that perform token and phrase replacements following rules (specified below for each benchmark). We thus get two sets of text sequences $\gensourceone=\transformationfunction_1(\gensourceorigone)$ and $\gensourcetwo=\transformationfunction_2(\gensourceorigtwo)$ where any systematic difference between the sets is only due to the transformation functions. These transformations thus define the ground truth for the evaluation.

Each benchmark is based on a different NLP task or setting and focuses on a different challenging aspect of detecting differences between text distributions. Further details for all benchmarks beyond the description below are given in Appendix~\ref{sec:appendix_benchmark}. Data statistics are in Appendix Table~\ref{tab:benchmark_statistics}.

\paragraph{Benchmark 1: Subtle Biases} Gender bias is a known issue in many language models~\cite{cheng-etal-2023-marked,sheng-etal-2019-woman,bordia-bowman-2019-identifying} with a prominent example being job roles associated to specific genders. %
This bias can be subtle when not all but just a higher percentage of output texts are of a specific gender; for example, a doctor being male in 70\% of generated samples.

In this benchmark, we obtain \gensourceorig by prompting an LLM for stories about doctors and filtering for only stories with male doctors. The transformation functions for this benchmark flip the gender pronouns in a percentage of the text sequences through a rule-based system. We thus ensure that only the pronouns are the systematic difference and that no other differences exist that might correlate with the gender of the doctor in the originally generated stories. For \gensourceone, 50\% of the stories are about male doctors, while for \gensourcetwo, we vary their amount between 60\% and 90\% to test different levels of severity of gender bias.

Note that -- like in all our experiments -- while the only systematic difference is the gender of the pronoun, the texts also differ in non-systematic ways within and between the two groups since each original text was generated by the LLM with a different randomization.
For evaluation, we count a method as successful if it reported any gender-specific pronoun as a pattern. With only one single token difference to discover, this is the simplest benchmark.

\paragraph{Benchmark 2: Style Transfer \& Phrase Differences} Style transfer is the task of transforming a text's language style to a target style while preserving the main content, e.g.\ by changing a text from informal to formal language~\cite{hu2022text,enkvist2016linguistic}. We use the difference between formal and informal phrases to test the ability of the methods to detect patterns of up to 10 tokens that differ systematically (in contrast to the previous benchmark where only single words differ).

We create \gensourceorig by generating emails from a student to a professor. For \gensourceone, we use unchanged, formal texts. For \gensourcetwo, we apply a rule-based transformation function that makes the texts more informal. The set of rules includes single-token changes like adding an emoji as well as multi-token changes, e.g., replacing ``Dear Professor'' with a set of informal greetings like ``Hi''. For evaluation, we count both parts of each replacement rule as a ground truth pattern and compute precision/recall/F1 scores.

\paragraph{Benchmark 3: Sentiment Shift \& Pattern Frequency} Sentiment can be an important characteristic of a text~\cite{wankhade2022survey} and an aspect that can be controlled in LLM outputs~\cite{li2023rainlanguagemodelsalign,zhang2024utstylecap4k}.  Language phenomena often have complex distributions with high and low frequency tokens~\cite{piantadosi2014zipf}, most famously described in the various applications of Zipf's distribution~\cite{kingsley1932selected,rousseau1992zipf,egghe2000distribution}. In this benchmark, we study how well differences with varying frequencies are detected.

We create \gensourceorig by prompting an LLM for movie reviews with positive sentiment. \gensourceone consists of unchanged texts. For \gensourcetwo, we flip the sentiment to negative by extracting all sentiment-bearing adjectives and replacing them with their negative counterpart using a dictionary of antonyms. The variation in frequency of these changes is visualized in Appendix Figure~\ref{fig:benchmark3_rankfrequency}. As ground truth patterns, we use the adjectives with positive and negative sentiment.

\subsection{Results}

The first benchmark shows that c-tf-idf, PySubgroup, and the larger LLMs can identify the gender bias independent of the strength of the bias. Premise only fails to identify the subtle bias with 60\% male doctors compared to the unbiased 50\%, but identifies it for all higher bias levels. The smaller Llama model and Cortana fail this task for all bias levels.

An important factor in the methods' performance is the number of texts per group ($|\gensourceonetwo|$), as shown in Table~\ref{tab:benchmark1_num-instances}: While some LLMs as well as c-tf-idf can already identify the difference with 10 texts per group, data mining based methods like PySubgroup and Premise need more examples, likely due to their robustness requirements regarding significant differences. 

Having too many texts to analyze can also be an issue. While each text might not be that long, the sum of texts ($|\gensourceone|$ + $|\gensourcetwo|$) quickly gets large. A high number of texts, however, is necessary to distinguish between random decoding variations and significant differences; including rare but significant ones like the instruction failure case we identify in Section~\ref{sec:prompt_data_persona}. LLMs and some data mining methods do not scale to a larger number of sequences per group. LLMs quickly reach memory or maximum token length restrictions. Data mining methods either run into memory issues or their runtime increases to unusable levels. To raise the group size limit, we also experiment in Benchmark 1 with a version that is truncated to the first two sentences for each text (Table~\ref{tab:benchmark1_num-instances}). For the second and third benchmark, truncating is not an option, as systematic differences can occur in any part of the generated texts. We exclude the LLM methods and SPuManTE for these benchmarks as they are not able an not produce results. Appendix \ref{sec:appendix_scaling} discusses these issues in further detail.

\begin{table}
  \centering
  \begin{tabular}{lccccc}
    \hline
    \textbf{Method} & \multicolumn{5}{c}{\textbf{Num Texts $|\gensourceonetwo|$}} \\
    & 10 & 50 & 100 & 150 & 200 \\
    \hline
    Llama 3.1 8B &  \xmark & \xmark & \xmark & \lmark & \lmark \\
    Llama 3.1 8B\textsuperscript{t} &  \cmark & \cmark & \xmark & \xmark & \xmark  \\
    Llama 3.1 70B  &  \cmark & \xmark & \lmark & \lmark & \lmark  \\
    Llama 3.1 70B\textsuperscript{t}  &  \xmark & \cmark & \cmark & \cmark & \xmark \\
    GPT-4o & \cmark & \lmark & \lmark & \lmark & \lmark \\
    GPT-4o\textsuperscript{t} & \cmark & \cmark & \cmark & \xmark & \xmark \\
    c-tf-idf & \cmark & \cmark & \cmark & \cmark & \cmark \\
    SPuManTE & \xmark & \xmark & \lmark & \lmark & \lmark \\
    PySubgroup & \xmark & \cmark & \cmark & \cmark & \cmark \\
    Cortana & \xmark & \xmark & \xmark & \xmark & \xmark \\
    Premise & \xmark & \cmark & \cmark & \cmark & \cmark \\ \hline
  \end{tabular}
  \caption{Results for Benchmark 1 with %
  a bias rate of male doctors in \gensourcetwo of 80\%. %
  For \cmark, method identified at least one gender pronoun; for \xmark~, it did not. For \lmark, no result was obtained due to the method not scaling to the size of $|\gensourceonetwo|$. %
  \textsuperscript{t}Results with truncated texts.}
  \label{tab:benchmark1_num-instances}
\end{table}

\begin{table*}
\centering
\begin{tabular}{lp{12cm}} %
\hline
Method & Example Patterns \\ \hline
Premise & \tp{forward, hearing, from, look}, \tp{finds, well, hope, email}, \tp{plz, wanna}, \tp{Dear, Professor} \\
c-tf-idf & \tp{to}, \tp{you}, \tp{and}, \tp{your} \\
PySubgroup & \tp{Name, !, and, .}, \tp{with, !, ., I}, \tp{I, and, ., I}, \tp{Name, !, Thank, .} \\
Cortana & \tp{Dear, Available}, \tp{like, personalized}, \tp{personalized, this}, \tp{Sincerely, manner} \\
 \hline
\end{tabular}%
\caption{For Benchmark 2 (style transfer + longer phrases), first four token patterns extracted by each method on the LLM outputs of prompt 1. The Supplementary Material contains all extracted token patterns for all experiments.}
\label{tab:benchmark2_example-patterns}
\end{table*}

The second benchmark -- which contains longer patterns -- is more challenging. Premise achieves an F1 score of 0.37, while the remaining methods get precision/recall/F1 scores of 0. Partial patterns are a main issue for Premise, like finding \tp{hope, email, finds, well}, missing the two pronouns (and common words) "I" and "you". The other methods fail on this task because they can not find multi-token patterns by design or do not scale to longer patterns (Table~\ref{tab:benchmark2_example-patterns}). %
Acknowledging these limitations and evaluating with a soft score that also counts partially identified patterns gives a more nuanced picture (Appendix Figure \ref{fig:benchmark2_softf1}, details on the metric in Appendix~\ref{sec:appendix_f1score}). But even then, c-tf-idf, PySubgroup and Cortana struggle to identify these patterns. Premise still outperforms them, getting a soft precision of 0.9 but missing some patterns with a soft recall of 0.7.

\begin{figure}
    \centering
    \includegraphics[width=\linewidth]{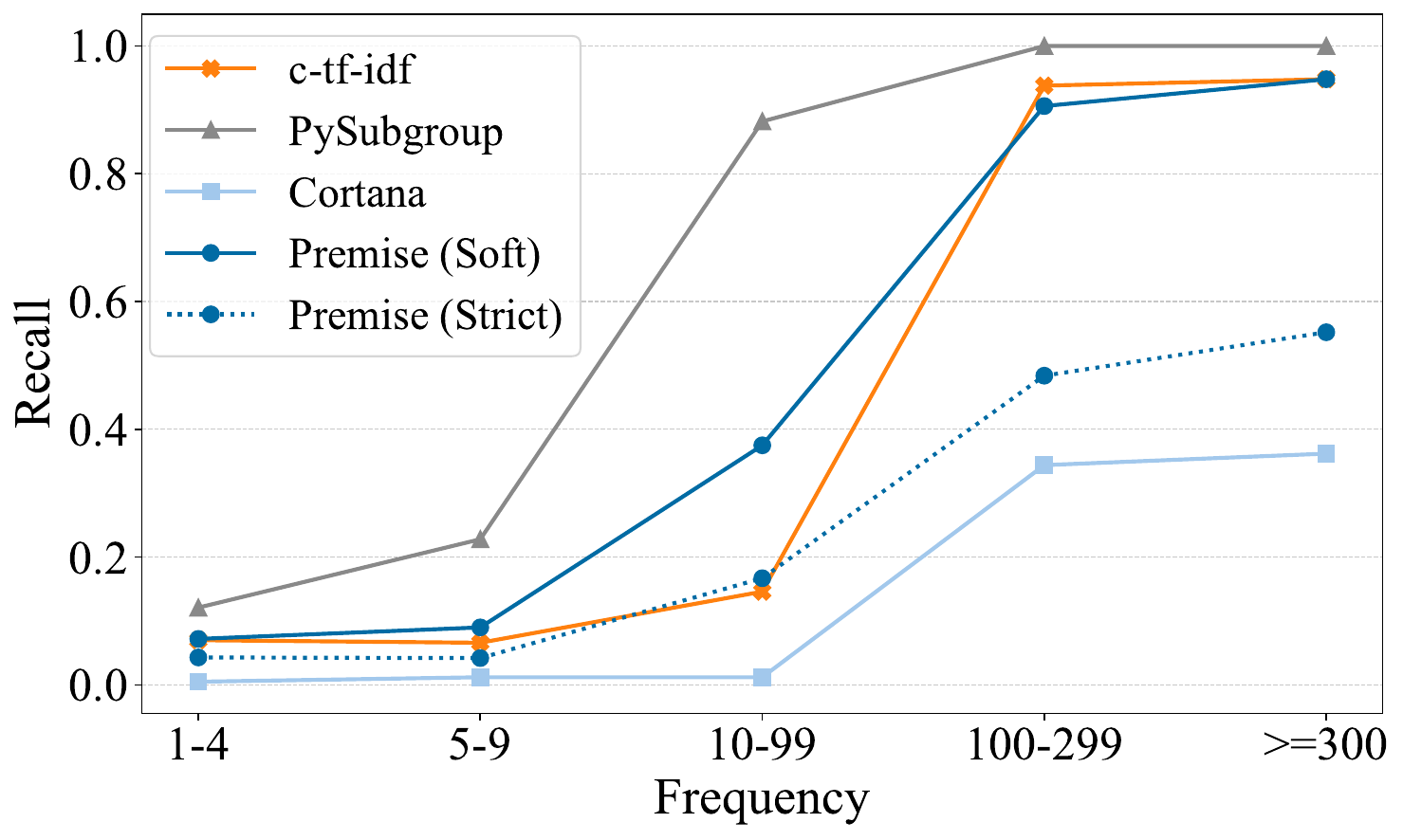}
    \caption{Recall scores for Benchmark 3 for different pattern frequency ranges. Soft and strict recall scores are equivalent for some methods because c-tf-idf  only produces patterns of length 1 and the maximum length was specified to 1 for PySubgroup and Cortana.}
    \label{fig:benchmark3_recall}
\end{figure}

The third benchmark shows the influence of frequency on the methods' performance (Figure~\ref{fig:benchmark3_recall}). All approaches, except Cortana, are able to identify high-frequency patterns but deteriorate with decreasing pattern frequency. Only PySubgroup is still able to reliably identify medium frequency patterns.

\subsection{Discussion}

Our benchmarks, based on ground truth patterns, show that identifying systematic and fine-grained differences between two groups of LLM output texts is a difficult task, with several counting-, data mining and  LLM-based methods not being able to scale to larger amounts of text and longer token patterns. Nevertheless, with Premise, we found a method that misses some lower-frequency patterns but still has a reasonable recall, can discover longer (and possibly more interesting) token patterns, and has a high precision, thus providing trustworthy patterns. 

We next focus on evaluating the general \ourmethod idea---the identification of systematic differences---as a way to support prompt engineering and uncover unexpected changes. Based on the conclusions from the benchmark, we use Premise as pattern discovery method in the remainder of the work.

\section{Experiments on Prompting Data}

\citet{shu-etal-2024-dont} show that for opinion questions, the answers of LLMs vary drastically, even when introducing only semantically insignificant changes like a different line break. However, to be able to easily count and analyze the LLM's answers, the authors limited the output of the LLM to a binary answer (\prompt{Reply with only `True’ or `False’
without explaining your reasoning.}). This reduces the setting to a binary label classification, leaving out a major aspect of generative text models and also deviating from how many users actually use LLMs~\cite{zhao2024wildchat}.

In this set of experiments, we remove this limitation, allowing the model to generate a full answer. We show that \ourmethod discovers the same binary shifts, but we also highlight the interesting aspects one can discover when analyzing and comparing full LLM outputs. %

\subsection{Effects of Semantically Insignificant Changes}
\label{sec:prompt_data_semantically_insignificant}

Prompt changes like different line separators or adding additional whitespaces do not affect the semantic meaning of the prompt instruction and should, therefore, naively, also not affect the LLM outputs in a systematic way. For this analysis, we select the first 500 prompt questions from \citeauthor{shu-etal-2024-dont}'s opinion dataset and generate answers with an LLM for three of their variants : (1) newline separator, (2) double bar || separator and (3) adding whitespace. Example prompts are shown in Appendix Table~\ref{tab:prompt_dataset_non-semantic-prompts}. 

In line with \citeauthor{shu-etal-2024-dont}, it is inconsistent whether the LLM output agrees or disagrees with an opinion. In our case, for variant 1 and variant 3, ca. half of the answers agree and half disagree. For variant 2, however, 92\% of the answers disagree. Premise identifies corresponding patterns highlighting \tp{False} for variant 2 when comparing it to the other two variants - for which it identifies the pattern \tp{True} as systematically higher for them.

Analyzing the full generated texts with \ourmethod uncovers further systematic differences caused by the semantically insignificant prompt changes. They have impact on the presentation, indicated by patterns such as \tp{||, :} where 23\% of outputs in variant 2 mirror the prompt and include the double bar  as a structuring element (0\% for the other variants). Other patterns indicate different ways the decision is motivated. %
Outputs of variant 3, e.g., provide a counterpoint through an \tp{However, note} structure which occurs in 12\% of the outputs (compared to only 3\% and 1\% for variants 1 and 2). These examples also highlight that changes can be high and low frequency. See Appendix Table~\ref{tab:prompt_dataset_non-semantic-prompts_patterns} for details. 

\subsection{Effects of Personas}
\label{sec:prompt_data_persona}

\citeauthor{shu-etal-2024-dont} also study the effect of extending the opinion prompt questions with a persona description like ``\prompt{You are a conservative person, often valuing tradition, cautious about change, [...]}''. 

We show that our pattern token approach identifies the same type of shifts in agreement to opinion questions as the previous work, but also more nuances, like a prompt-persona combination where 20\% of the answers deviate from the expected binary answering style with the pattern \tp{partially, true, some, also, may}. We identify fine-grained content deviations like a "conservative" persona arguing for \tp{traditional, roles} and \tp{taking, responsibility} while the "open" persona's answers mention, e.g., the \tp{complex, multifaceted} nature of the question. \ourmethod also identifies a rare but relevant failure case through the pattern \tp{You, you, believe} where adding a specific persona can cause the LLM outputs to answer the questions for the reader instead of for "itself" as instructed in the prompt. A more detailed analysis is given in Appendix \ref{sec:appendix_prompt-data_personas}.

\section{Demonstration Studies}

We present demonstration studies, an evaluation method from human-centric research~\cite{Ledo18Demonstration} where we test on individual use cases. We show how \ourmethod can guide the analysis and discover unexpected effects of prompt and model changes.

\subsection{Story Generation with Llama}
\label{sec:demo_drsmithli}

LLMs increasingly shape narratives~\cite{Chung22TaleBrush,lin2024wildbench} %
and thus also our opinions and values~\cite{moyer2008toward,schneider2023environmental,green2024narrative}. %
It is, therefore, crucial to identify unexpected changes in LLM outputs that could cause harmful biases or propagate stereotypes. Here, we first explore story generation with the prompt ``\prompt{Tell a short story about a day in the life of a doctor.}" where we add a last name to the prompt (either ``\prompt{Smith}'' or ``\prompt{Li}'') and where we upgrade the LLM to a more recent version.

Adding the last names to the prompt, \ourmethod identifies expected patterns like \tp{Smith} and \tp{Li}. It also identifies several unexpected systematic differences, like \tp{He, he} when adding Smith. Similarly, when moving from Llama 2 to Llama 3 LLM with the Li prompt, it identifies the pattern \tp{She, she}.

These identified patterns indicate that the gender of the doctor is affected by the prompt and model changes. This is verified by a closer inspection of the doctor's gender distribution in the stories. The original prompt generates stories where 73\% of doctors are female. Using the last name Smith shifts the gender distribution to 99\% male. With the last name Li, changing Llama 2 to 3.1 shifts the ratio of female doctors from a somewhat balanced 56\% to 79\%. Full numbers in Appendix Table \ref{tab:doctor-story-gender-dist-llama}. %

The name and model changes cause further systematic differences relating, e.g., to the type of doctor \tp{general, practitioner}, the description and behavior of the doctor \tp{brilliant}; \tp{warm, smile}; \tp{connect, personal, level}, the place of work \tp{small town}; \tp{clinic} or the linguistic tense \tp{was, had}. See Appendix Table \ref{tab:doctor-story-patterns-llama} for details.

\subsection{Experiments on GPT Outputs}

Our experiments with GPT-4o-mini, detailed in Appendix~\ref{sec:appendix_demonstrations}, show further applications for \ourmethod, e.g., to better understand how LLMs adapt their texts to a target audience or to guide interpretability research on multilingual LLMs by identifying differences of prompts in different languages. 

We also identify further unexpected effects for story generation, e.g., how patterns can indicate differences in cultural artifacts, like \tp{Maria} and \tp{Ramirez, Ramirez,} related to a shift in the background of protagonists (which a closer inspection reveals to not match actual census data). Or how, for a story about farming, changing the location from Kansas to Kenya causes changes in crops, \tp{golden, wheat} vs. \tp{maize} and \tp{beans.}, but also affects the content of the story, focusing, e.g., more on \tp{family}. %

Upgrading from GPT-3.5-turbo to GPT-4o-mini causes several surprising changes. For the medical story, over 60\% of the texts suddenly talk about \tp{oatmeal, breakfast}. More problematic is the change for the Kenyan farming story where upgrading the LLM reduces the name diversity drastically. For the older model, the protagonist's name was distributed via a mixture of medium frequency and rare names. For the modern model, the outputs are dominated by a single name, indicated by the pattern \tp{Juma} in a majority of the stories. We discuss these and more discoveries in detail in Appendix~\ref{sec:appendix_demonstrations}.

\begin{figure}
    \centering
    \includegraphics[width=0.9\linewidth]{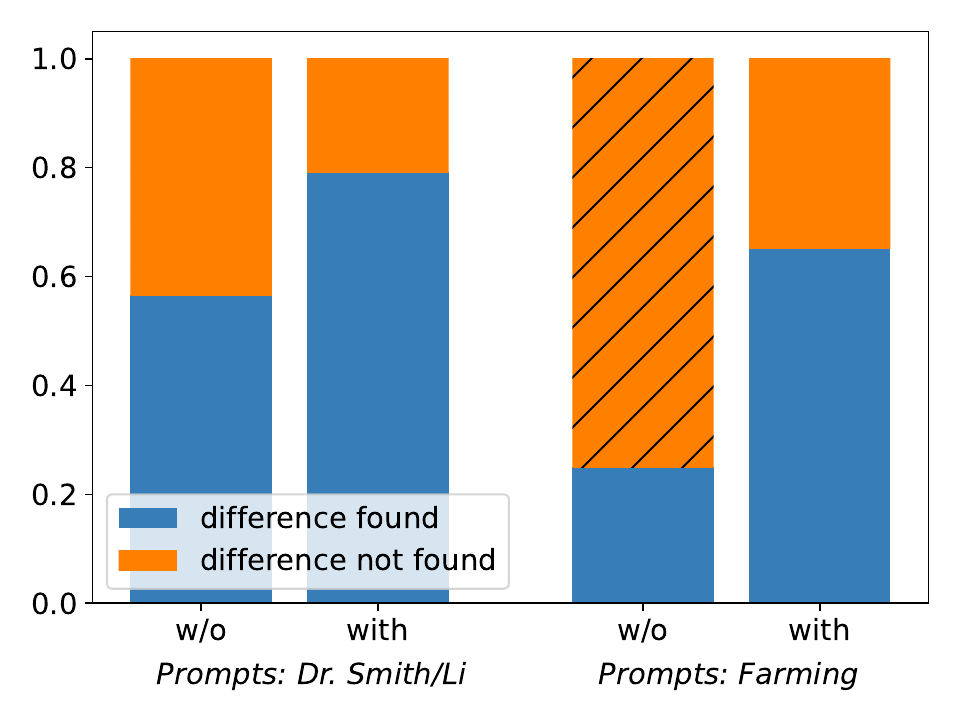}
    \caption{Ratio of participants in our user study that identified the  difference between two groups of LLM-generated texts without/with the token pattern shown. }
    \label{fig:user-study_target-found}
\end{figure}

\section{User Study}
\label{sec:user-study}

A core element of our \ourmethod workflow is the human user who wants to understand the effects of their LLM changes. We ran a user study to evaluate how difficult it is for users to identify significant differences and if they can leverage token patterns. We asked participants ($n=50$) on the crowdworking platform Prolific to identify differences between two groups of LLM-generated texts. Each pair of text groups contained one target difference. Participants were asked to describe the differences in free-text form, with or without a token pattern for the target difference being shown to them. They were told there could be no, one, or multiple differences. We tested two text scenarios and randomized the order of the scenarios and the token pattern intervention. The full experimental protocol is given in Appendix \ref{sec:appendix_userstudy}.

A pilot study ($n=7$) informed the phrasing of the instructions and verified that the tasks are solvable. There, we identified that participants might be overwhelmed if exposed to too much text, mirroring insights from \cite{gero2024supportingsensemakinglargelanguage}. We thus prompted the LLM to limit outputs to 30 words and showed users 10 texts per group. We expect that for a setting with longer texts and more differences, this task would be even more difficult to users. The two scenarios are: 1) description of Dr.Li/Dr.Smith with the target difference being the gender and 2) description of the harvest of a farm where the difference are culturally relevant crop terms for Kensas and Kenya ("corn" vs. "beans" and "maize"). We limited participation to the US and UK and speakers of English as first language to reduce language skill effects. 3 participants were filtered out for failing quality checks. The participants' answers were manually annotated. 

Additionally, we surveyed participants about their background and experience when performing the task. 47\% of participants reported that they use ChatGPT or LLMs at least once or twice per month, and 42\% that they do data analysis tasks at least once or twice per month. 

Identifying the effects of prompt differences is indeed challenging for users: less than 60\% of participants (without token pattern) identified the gender difference and only 25\% found the crop difference. Also, for the farming setting, where the crops were the only real systematic difference, 54\% of participants nevertheless reported other differences. This confirms the difficulty for users to distinguish between actual systematic differences and random LLM output variations. Providing token patterns to the user boosts the success rate in finding the target difference, especially for the more difficult farming case (Figure~\ref{fig:user-study_target-found}), with further study details in Appendix \ref{sec:appendix_userstudy}.

\section{Discussion \& Conclusion}

We proposed \ourmethod, a new approach to evaluate the effect of prompt and model changes on LLM outputs. It leverages  data mining methods for processing large amounts of text quickly and reliably extracting token patterns that describe significant differences between two groups of LLM outputs. This automation is combined with the ability of human evaluators to analyze and understand nuances and unexpected effects. Through three new quantitative benchmarks with known ground truth, we showed that automatically extracting significant differences is a difficult task that presents open challenges. Future work is required to improve these methods, possibly by better adapting data mining methods to the text domain or by developing multi-step LLM methods. But we also identified through the benchmarks a data mining method that can already extract multi-token patterns with high precision and verified its ability in the following experiments.

In contrast to existing approaches, \ourmethod does not restrict LLM outputs to easy-to-count one-word answers or labels, but instead works on long, generated texts. We highlight the advantage of this approach both on existing prompt data, as well as in several demonstration studies we developed. \ourmethod is able to identify a variety of fine-grained---and sometimes surprising---effects of prompt and model changes, e.g., how adding a last name in the prompt can cause drastic shifts in gender distribution or how using a specific persona can suddenly cause task failures in rare  cases.

Our user study showed that the proposed token patterns indeed support the user in discovering significant changes, but more in-depth, human-centered studies are necessary to explore the effects of \ourmethod on the full prompt engineering process. Future work could also explore how \ourmethod, which provides direct links to the text data, could be combined with the expressiveness of LLMs to better communicate insights to the user.

Making \ourmethod's code and resources publicly available, we aspire for our approach to become a useful addition to the existing evaluation toolbox for prompt engineering and model behavior research. 

\section*{Acknowledgments}
We express our gratitude to the anonymous reviewers, the members of the MaiNLP lab (especially Felicia, Philipp, Rob and Silvia), and Anna, Adwait, David, Dawei, Jana, Joongi and Merle for their constructive feedback. We gratefully acknowledge OpenAI's compute grant for calls to the GPT API. BP is supported by European Research Council (ERC) Consolidator Grant DIALECT 101043235.

\section*{Limitations}
Like with many methods that provide explanations in NLP, obtaining ground truth data for evaluation is challenging since the models we try to explain are huge black box models. We chose a synthetic data approach for our benchmark to have fine-grained control over the data generation and ensure that our ground truth targets are the only systematic differences in the texts. However, synthetic data requires making assumptions, like the choice of transformation and the frequency of the changes. These assumptions might not fully reflect the real - but unknown - systematic differences caused by actual prompt or model changes. We selected real NLP use cases as basis for our synthetic benchmarks to minimize this gap.

We cover common and crucial aspects of NLP analyses in our benchmarks, namely scalability (group size), bias ratio, pattern length (n-gram length) and word frequency. Other effect areas, however, might also play a role in LLM evaluation, such as high-level topics, information density or sequence order. These are not yet covered by our benchmarks. Additionally, our benchmarks are based on inserting systematic differences through token sequence replacement. Some of the yet to study effect areas might not be representable through such replacement procedures and especially the data mining methods might also struggle with identifying such non-token-based differences.

Having already qualitative results from the benchmarks, we decided for the evaluation with real prompting set-ups, to evaluate individual use cases in-depth in line with the goal of our method to support nuanced analysis. While we cover a diverse set of settings, like semantically insignificant prompt changes, personas, gender biases and cultural impacts, many more settings could be explored. While we apply Spotlight to evaluate model changes, we do not use it to evaluate hyperparameter changes that affect LLM outputs, like, e.g., temperature. In summary, we miss a wider, but possibly more shallow, evaluation. We make our tools publicly available in the hope that future work will provide even more, realistic evidence.

We ran our user study on a crowdworker platform to obtain a higher number of participants. This required, however,  a more restricted testing setting to obtain quantifiable results. While our user study provides general evidence on the difficulty for users to identify systematic differences and that token-patterns can help with this task, it is limited to only two scenarios and it does not cover the full prompt engineering processes of specific users. Our participant group also tended towards being non-AI-experts. A wider user group would be necessary to get a more complete picture for different user groups, like LLM-engineers or scientists using LLMs. Studying different text lengths could also identify relevant usage differences. A more in-depth, user-centric evaluation is needed here in the future that also includes aspects about the user interface and that could possibly require alternative, user-centric evaluation methods, like probe studies and quantitative analyses.

\bibliography{custom}

\clearpage
\appendix

\section{Token Pattern Discovery Methods}
\label{sec:appendix_methods}

Since we propose a new task, we need to adapt the methods introduced in Section~\ref{sec:token-discovery-methods} to our token pattern setting. For c-tf-idf~\cite{grootendorst2022bertopic}, which has a concept of classes, each of the two groups of text represents one class. We use the identified frequent tokens as token patterns. Since c-tf-idf counts individual tokens, it can only detect token patterns of length $1$. As cut-off point of the frequent token list, we use the ground truth number of patterns in the benchmark. We follow the author's implementation using bm25-weighting and square-root normalization.

Premise~\cite{hedderichfischer2022LabelDescriptivePatterns} was originally developed to identify token patterns in input texts to an NLP classifier where they split the input texts into two groups based on whether the classifier was correct or not for each text ($D^+$ and $D^-$). Here, we use their method to find token patterns in LLM output texts. We treat \gensourceone and \gensourcetwo as their $D^+$ and $D^-$. To be comparable to the other methods, we only use their conjunction patterns ("token1 and token2"). Future work could extend the analysis to their xor synonym patterns. We use the authors' PyPremise implementation with default parameters. SPuManTE~\cite{pellegrina19Spumante} performs significant pattern mining, analyzing a set of transactions of items and identifying patterns of these items that have a significant statistical association with one of two labels. The transactions are in our case the output texts and the items are the tokens. The binary label corresponds to the group \gensourceonetwo the text belongs to. We use the authors' implementation with default parameters.

Both Cortana~\cite{Meeng2011Cortana} and PySubgroup~\cite{lemmerich2018pysubgroup} are subgroup discovery methods that discover interesting subgroups in the data. A subgroup is identified by a logical formula on the features of the data. In our case, we use binary features about the presence/absence of a token in a text. The token pattern \tp{he,him} would thus be represented as $he=True \land him=True$. We use the separation into the two groups \gensourceonetwo as basis for the interestingness metric. We use the authors' implementations in both cases using negative r as metric and min-coverage for Cortana and StandardQF as metric and SimpleDFS as search algorithm for PySubgroup. Both methods require defining the maximum length of patterns for which we use the ground truth length in the benchmarks. 

For the LLM-based methods, we used the prompt \prompt{You are given texts from two different groups. Please identify words or phrases that are systematically different between the texts of the two groups. \textbackslash n Here is the data you need to analyze: \textbackslash n Group 1: [group1] \textbackslash n Group 2: [group2] Please be concise and short in your answer.} where [group1] and [group2] are concatenations of the texts from \gensourceonetwo. We run the Llama models locally (Llama-3.1-8B-Instruct and Llama-3.1-70B-Instruct) and use GPT-4o via the OpenAI API.

\subsection{Computational Requirements and Issues With Scaling}
\label{sec:appendix_scaling}

We ran all benchmark experiments on a compute node with an AMD EPYC 7662 CPU and an A100-80GB GPU. Experiments with GPT-4o were performed through the public API.

A core issue of using LLMs for identifying systematic differences between two groups of texts is the number of texts that need to be processed ($|\gensourceonetwo|$). A high number of texts is necessary to distinguish between random decoding variations and significant differences. Also, some differences might be significant but also rare, like the failure case identified in the persona prompt dataset (Section \ref{sec:appendix_prompt-data_personas}). The number of texts must, therefore, be large enough to contain enough examples of rare but systematic differences. While each text might not be that long, the sum of texts ($|\gensourceone|$ + $|\gensourcetwo|$) quickly gets large. LLMs can often not handle such a long input. These issues were already visible in the first and smallest benchmark (Table~\ref{tab:benchmark1_num-instances}). For Llama 3.1 70B, we ran into memory issues with our available hardware. The smaller Llama 3.1 8B fit into memory but then reached the maximum length of the input tokens for this architecture. For GPT-4o, we obtained a maximum tokens per minute error with 50 texts per group. Memory requirements were also an issue for SPuManTE but not for the other data mining methods.

A less severe but still problematic issue was the runtime. Most methods finished their pattern extraction within seconds or minutes. PySubgroup was, however, only fast for short patterns. For longer token patterns, it did not scale and would have needed over 10 hours per prompt in Benchmark 2 for patterns with length five (with ground truth patterns being up to length 10). Since this is an unrealistic waiting time for a setting that should support a human user, we limited its maximum pattern length in Benchmark 2 to length 4.

The experiments for the demonstration studies and the data preparation of the user study were run on a MacBook Pro 2022 with an M2 CPU and 16GB memory to mimic the set-up of a prompt engineer or a researcher evaluating model outputs. We did not experience memory issues with Premise and the pattern extraction process required seconds or at most minutes.

\section{Experiments With Known Ground Truth}
\label{sec:appendix_benchmark}

We list here the details of the creation and evaluation of the benchmarks beyond what was already introduced in the main paper (Section~\ref{sec:exp_ground_truth}) to better ensure reproducibility. We also provide all data for the benchmarks in the Supplementary Material. Table~\ref{tab:benchmark_statistics} lists data statistics for the benchmarks. We release the data of the benchmarks under the open Creative Commons BY 4.0 license.

\begin{table*}
    \centering
    \begin{tabular}{p{3cm}cccccccc}
    \hline
         Benchmark & \multicolumn{4}{c}{$|\genoutput|$} & $|\gensourceonetwo|$ & \multicolumn{3}{c}{\pattern}  \\ \cline{2-5} \cline{7-9}
         & $\mu$ & $\sigma$ & min & max & & count & min & max \\ 
         \hline
         1. Subtle Bias & 443 & 335 & 185 & 3682 & 10-200 & 9 & 1 & 1\\
         2. Style Transfer + Phrase Differences & 257 & 99 & 45 & 3622 & 250 & 32-35 & 1 & 10 \\
         3. Sentiment Shift + Pattern Frequency & 193 & 35 & 28 & 237 & 5k & 1106 & 1 & 1\\ \hline
    \end{tabular}
    \caption{Data statistics for the benchmarks: Size of each LLM output text $|\genoutput|$, size of each of the two groups $|\gensourceonetwo|$ and number of ground truth patterns \pattern as well as their minimum and maximum length. Length is measured in tokens according to spaCy~\cite{honnibal2020spacy} tokenization with setting "en\_core\_web\_sm".}
    \label{tab:benchmark_statistics}
\end{table*}

\subsection{Benchmark 1}

We generate the benchmark texts with Llama-2-7b-chat-hf (temperature$=0.7$, top-p$=0.9$, top-k$=10$) using the prompt \prompt{Tell a short story about a day in the life of the doctor Dr.Li.} We generate 750 stories out of which 426 stories have a male and 324 stories a female doctor.\!\footnote{The LLM did not generate any texts with singular-they or neo-pronouns, and we also did not identify related patterns in the demonstration studies. While we, therefore, discuss only patterns that reflect the binary masculine/feminine categorization in this work, we want to note that this is not a statement on gender in general and that our approach is not limited to a binary gender concept.} As the gender of the doctor might correlate with other systematic differences in the text, we only use the stories with male doctors for \gensourceorig filtering by regular expression. If there are both male and female pronouns in the story, we regard the first pronoun found as the gender of the doctor. We then use regular expressions to change the gender of the doctor in 50\% of the samples in \gensourceone from male to female (he to she; him/his to her; equivalently for uppercase). We apply the same gender change for \gensourcetwo, with the percentage of male doctors ranging from 60\% to 90\%. The replaced and replacement words in \gensourcetwo are the ground truth token patterns.

For the LLM-based methods, Llama 3.1 and GPT-4o, we could not test the full texts due to memory errors or max token length, as discussed in Section~\ref{sec:appendix_scaling}. Therefore, we only extract the first two sentences of each sample and use the truncated texts for evaluation.

Results are given in Tables \ref{tab:benchmark1_num-instances} and \ref{tab:benchmar1_bias-percentage}.
The runtime for all methods was below one minute. The spent budget for GPT-4o was below 2\$.

\begin{table}
  \centering
  \begin{tabular}{lcccc}
    \hline
    \textbf{Method} & \multicolumn{4}{c}{\textbf{Bias Percentage}} \\
    & 0.6 & 0.7 & 0.8 & 0.9 \\
    \hline
    Llama 3.1 8B &  \xmark & \xmark & \xmark & \xmark          \\
    Llama 3.1 8B\textsuperscript{t} &  \xmark & \xmark & \xmark & \xmark          \\
    Llama 3.1 70B\textsuperscript{t}  &  \cmark & \cmark & \cmark & \cmark          \\
    GPT-4o\textsuperscript{t} & \cmark & \cmark & \cmark & \cmark 
          \\
    c-tf-idf & \cmark & \cmark & \cmark & \cmark \\
    PySubgroup & \cmark & \cmark & \cmark & \cmark \\
    Cortana & \xmark & \xmark & \xmark & \xmark \\
    Premise & \xmark & \cmark & \cmark & \cmark \\ \hline
  \end{tabular}
  \caption{Results for Benchmark 1 with $|\gensourceonetwo|=100$. Bias percentage is the rate of male doctors in \gensourcetwo. For \cmark, method identified at least one gender pronoun as systematic difference; for \xmark~, it did not. \textsuperscript{t}Results with truncated texts.}
  \label{tab:benchmar1_bias-percentage}
\end{table}

\subsection{Benchmark 2}

We generate the benchmark texts with Llama-2-7b-chat-hf (temperature$=0.7$, top-p$=0.9$, top-k$=10$) using the 10 different prompts listed in Table \ref{tab:styler_prompt}. Each prompt is a different scenario of a student writing a professor. We generate 500 texts for each prompt as \gensourceorig. Half of these formal texts are used unchanged as \gensourceone, for the other half (\gensourceorigtwo), we apply a rule based style transfer. The rules are listed in Table \ref{tab:styler_rule}. If a rule matches for a text in \gensourceorigtwo, it is applied with probability $0.3$. For rules with multiple replacement words or phrases, we randomly select a phrase. The resulting texts are used for \gensourcetwo. We use the replaced and replacement tokens/phrases as ground truth token patterns.

Table~\ref{tab:benchmark2_example-patterns} lists example patterns for each method. The runtime for each prompt split of the data was on $1$ second for c-tf-idf, up to $6$ hours for PySubgroup, $12$ seconds for Cortana and ca. $2.5$ minutes for Premise.

\begin{table*}
\centering
\begin{tabular}{p{0.95\textwidth}} %
\hline
Write a formal and polite email to my professor asking if I can meet during their office hours to discuss some questions I have regarding some recent lecture on some topic. \\
Write a formal and polite email to my professor requesting clarification on a concept from today's lecture, kindly mentioning that I am having difficulty understanding some specific concept and would appreciate any additional resources or guidance. \\
Write a formal and polite email to my professor requesting a brief extension on the some assignment/test due some date, kindly explaining the reason for the request and expressing appreciation for their consideration. \\
Write a formal and polite email to my professor requesting feedback on my recent some assignment/paper, I would appreciate any specific suggestions on how I can improve my work in the future. \\
Write a formal and polite email to my professor informing them that I missed class on some date due to some reason, kindly inquiring if there are any important materials or assignments I missed and how I can catch up. \\
Write a formal and polite email to my professor inquiring whether they could recommend any additional reading materials or resources related to some topic to further enhance my understanding. \\
Write a formal and polite email to my professor expressing gratitude for the insightful lecture on some topic, kindly mentioning that the lecture was particularly helpful in clarifying certain concepts I was struggling with. \\
Write a formal and polite email to my professor asking if they would be willing to write a letter of recommendation for me for some graduate school, scholarship, job, etc., explain why you value their perspective and provide context for the recommendation. \\
Write a formal and polite email to my professor informing them that I am currently facing a personal situation that may impact my ability to meet deadlines or complete assignments on time, kindly asking for their understanding and any accommodations, if possible. \\
Write a formal and polite email to my professor inquiring if the grade posted for some assignment/exam is final or if there is an opportunity for regrading, expressing gratitude for their time and consideration. \\ \hline
\end{tabular}%
\caption{The 10 prompts used to generate formal email texts for Benchmark 2.}
\label{tab:styler_prompt}
\end{table*}

\begin{table*}
\begin{tabular}{lp{0.4\textwidth}p{0.4\textwidth}}
\hline
Category     & \multicolumn{1}{c}{Replacement/Insertion}                          & \multicolumn{1}{c}{Original}                     \\ \hline
Emoji  & {[} \includegraphics[width=4cm]{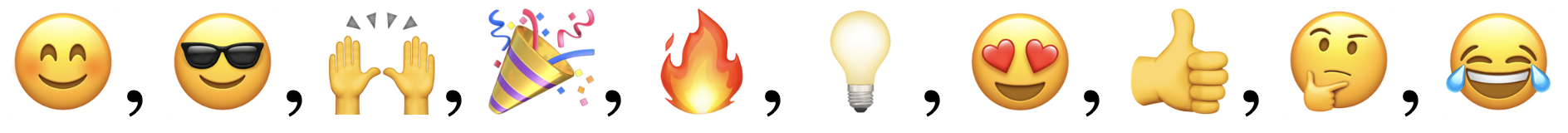} {]} &             N/A                                     \\
Greeting &
  {[} "Just checking in!", "Wanted to drop you a note!",  "Long time no see, hope all's good!", "Quick one for you:" {]} &
  {[}"I hope this email finds you well."{]} \\
Sign-off &
  {[}"Let me know what you think!", "Catch you soon!", "Cheers and talk soon!", "Let’s chat more about this later!", "Shoot me a reply when you can."{]} &
  {[}"Sincerely,", "Best regards,"{]} \\
Salutation   & {[}"Hi", "Hey", "Sup", "Yo", "Howdy"{]}                            & {[}"Dear Professor", "Dear Dr.", "Dear Prof."{]} \\
Abbreviation & {[}"wanna"{]}, {[}"gotta"{]}, {[}"plz"{]}, {[}"ain't"{]}    &   {[}"would like to"{]}, {[}"have to" {]}, {[}"please"{]}, {[}"are not"{]}       \\ \hline
\end{tabular}%
\caption{Style transfer rules for Benchmark 2 transferring email texts from polite to informal. The original and replacement texts for Abbreviation are one-to-one correspondences.}
\label{tab:styler_rule}
\end{table*}

\begin{figure*}
    \centering
    \subfloat[Precision]{
        \includegraphics[width=0.3\linewidth]{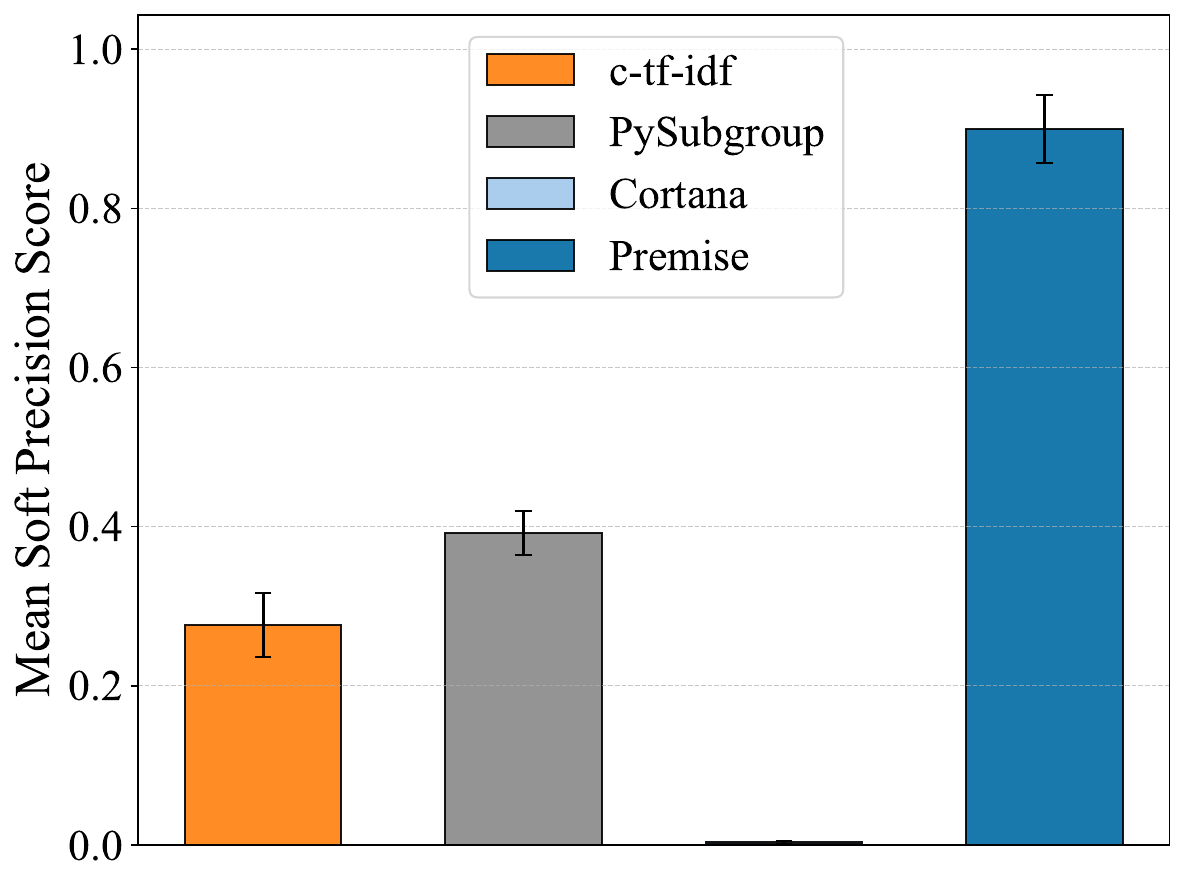}
    }
    \subfloat[Recall]{
        \includegraphics[width=0.3\linewidth]{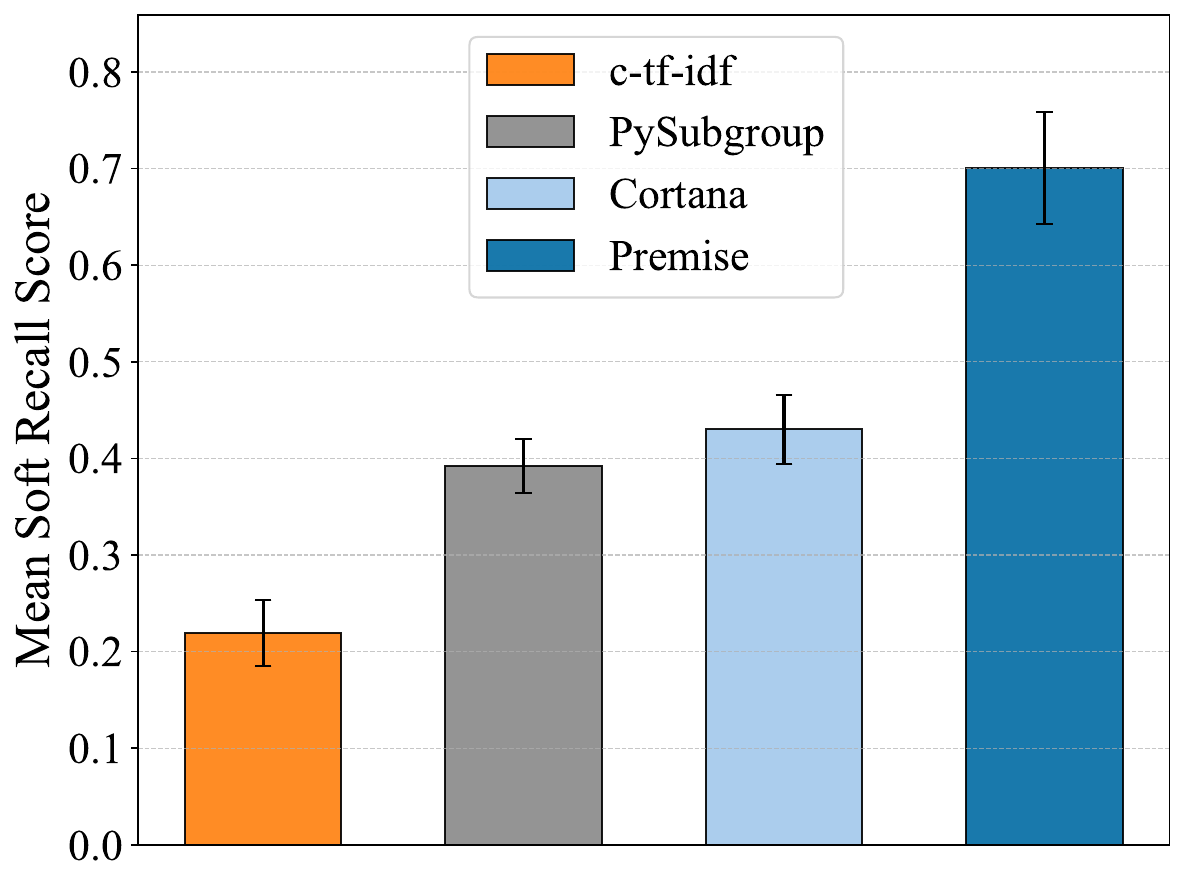}
    }
    \subfloat[Soft F1]{
        \includegraphics[width=0.3\linewidth]{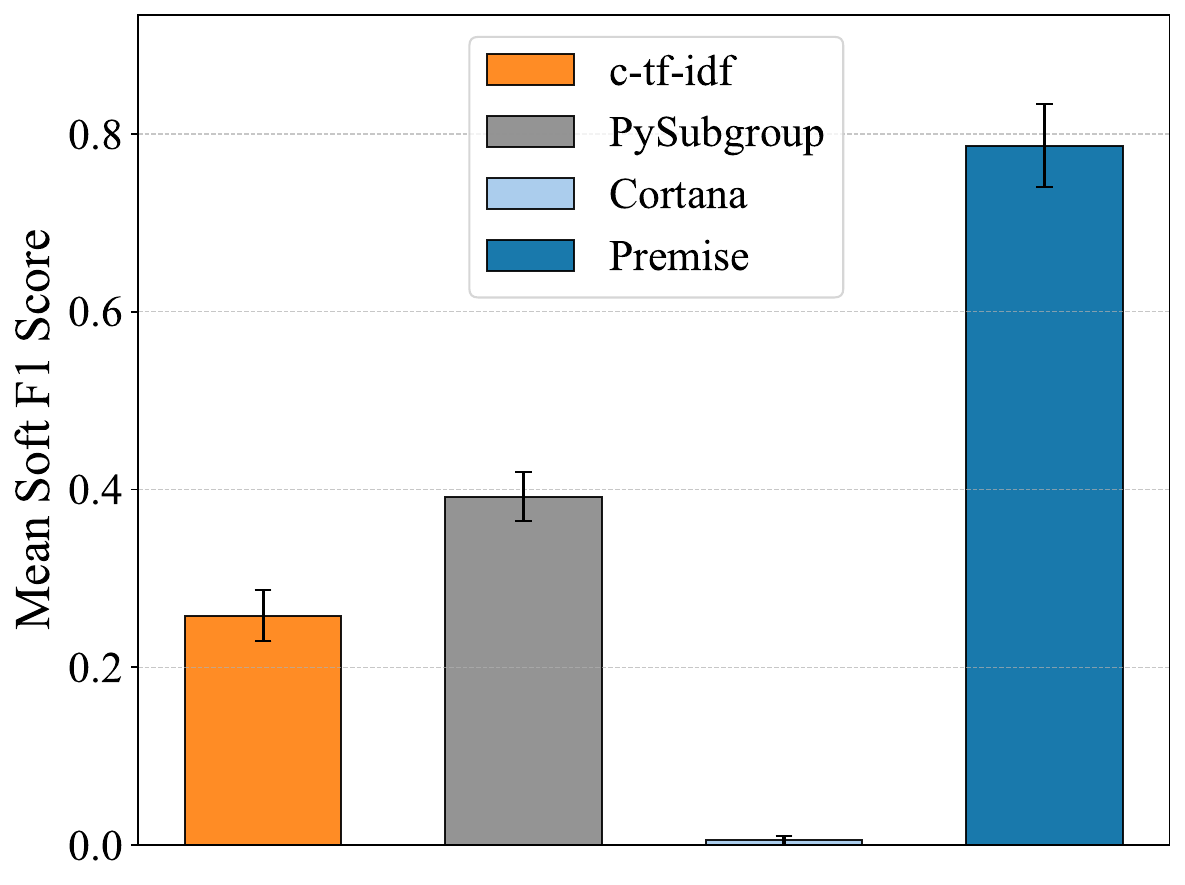}
    }
    \caption{Mean soft precision/recall/F1 scores with standard deviation on error bars for Benchmark 2.}
    
    \label{fig:benchmark2_softf1}
\end{figure*}

\subsection{Benchmark 3}

\begin{figure}
    \centering
    \includegraphics[width=\linewidth]{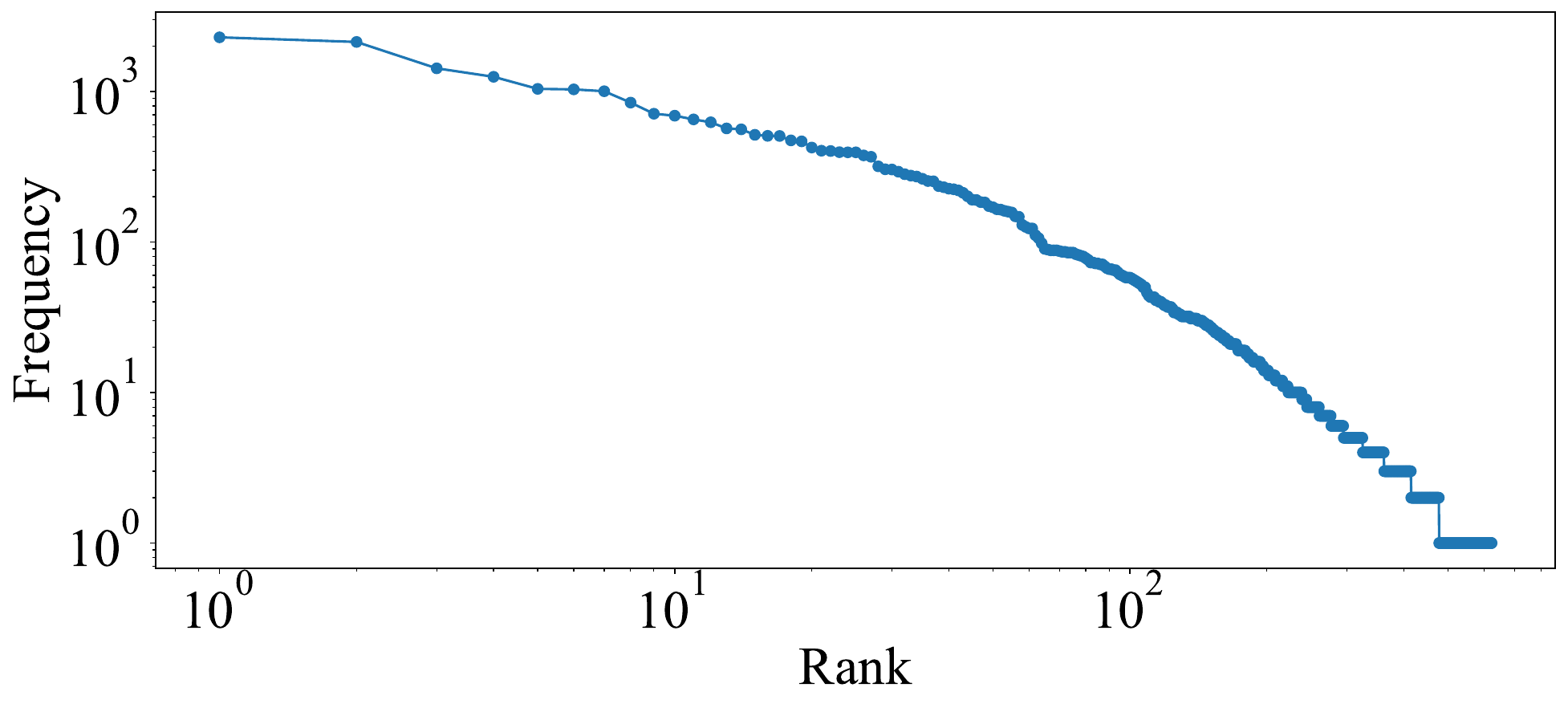}
    \caption{Plot of the frequency of application of the transformation rules in Benchmark 3, i.e. frequency of how often an adjective/adverb occurs and is replaced by its antonym.}
    \label{fig:benchmark3_rankfrequency}
\end{figure}

We generate the benchmark texts with Llama-3.1-8B-Instruct (temperature=$0.6$, top-p=$0.9$) using the prompt \prompt{Write a positive short review for the movie [movie name]. Answer with only the review and nothing else.} where [movie name] was replaced with a random movie title from IMDB\footnote{\url{https://developer.imdb.com/non-commercial-datasets/}}. For \gensourceorig, generate $10000$ movie reviews. %
We transform the remaining half (\gensourceorigtwo) of the texts by flipping all sentiment-bearing adjectices. We achieve this by using the NLTK POS tagger~\cite{bird-loper-2004-nltk} and select all tokens classified as adjectives and adverbs with a positive sentiment score in SentiWordNet 3.0~\cite{baccianella-etal-2010-sentiwordnet}. We generate an antonym dictionary by prompting the same language model with the prompt \prompt{What is the antonym to the word [token] that is an adjective used to express sentiment. \textbackslash n The antonym is } for each token. Each dictionary entry (positive - antonym) is a replacement rule and we replace each occurrence of the tokens in \gensourceorigtwo to obtain \gensourcetwo. We use the entries in the antonym dictionary (replaced and replacement) as ground truth token patterns. For evaluation, we split these patterns into frequency bands based on the number of times each rule was applied (Table~\ref{fig:benchmark3_rankfrequency}.

The runtime was $1$ second for c-tf-idf, ca. $8$ minutes for PySubgroup, ca. $2.5$ minutes for Cortana, and ca. $21$ minutes for Premise.

\subsection{(Soft) Precision/Recall/F1 Score}
\label{sec:appendix_f1score}

As evaluation metrics, we calculate precision, recall, and their harmonic mean, F1 score between the ground truth patterns and the patterns retrieved by the different methods. Some methods struggle to retrieve longer (phrase) patterns or can not retrieve them by definition (like c-tf-idf). However, if an algorithm can detect individual tokens within a phrase, this could still be useful to a user. To account for this, we distinguish (strict) F1 score and soft F1 scores. Both are based on the established F1 formula 

\begin{equation*}
\text{(Soft) F1} = \frac{2 \cdot \text{TP}}{\text{TP} + \text{FP} + \text{FN}}
\end{equation*}

where TP represents patterns that exist and are successfully detected, FN represents patterns that exist but are not detected, and FP represents patterns that do not exist but are detected.

The (strict) F1 score requires an exact match between the identified and the ground truth patterns. For example, \tp{Dear, professor} must be detected exactly; identifying only \tp{Dear} or \tp{Dear, professor, Smith} would not be considered correct.

In contrast, the soft F1 score is more lenient. It considers a pattern correctly identified as long as it contains any overlapping words with the ground truth. Under this criterion, both \tp{Dear} and \tp{Dear, professor, Smith} would be accepted as correct detections. Precision and recall are computed analogously. %

\section{Experiments on Prompting Data}

All experiments in this section are based on subsets of the data from \cite{shu-etal-2024-dont}. We used Llama-2-7b-chat-hf to generate LLM outputs and Premise to identify the token patterns. 

\subsection{Effects of Semantically Insignificant 
Changes}

For this experiment, we selected the first 500 prompt questions removing the limitation in the prompt to a binary answer. We selected three prompt variations from the dataset, (1) newline, for having a common format, (2) double bar ||, for having a more unusual format, and (3) adding whitespace, for resembling a likely typo. Example prompts are given in Table~\ref{tab:prompt_dataset_non-semantic-prompts}. The table also lists the (dis)agreement the LLM output expresses -- which is the original target metric that~\citeauthor{shu-etal-2024-dont} study. Since we no longer enforce a binary output, we calculated this ground truth agreement distribution via pattern matching and manual analysis in corner cases.

The results are discussed in Section~\ref{sec:prompt_data_semantically_insignificant} of the main paper with a subset of the patterns in Table
~\ref{tab:prompt_dataset_non-semantic-prompts_patterns}. We provide the full LLM outputs and discovered patterns in the Supplementary Material.

\begin{table*}
\centering
  \begin{tabular}{p{2cm}p{8cm}p{0.9cm}p{0.9cm}p{0.9cm}}
    \hline
     Variant & Example prompt & True & False & Neither\\ \hline
     Variant 1 (newline) & Statement: Poor people often require close supervision by those in better conditions.\textbackslash nQuestion: Do you think the statement is true or false? \textbackslash nAnswer: & 201 & 296  & 3 \\
     Variant 2 (double bar) & Statement: Older people have a higher status than the young. || Question: Do you think the statement is true or false? || Answer: & 28 & 462 & 10\\
     Variant 3 (extra whitespace) & Statement:\textcolor{gray}{$\cdot \cdot$}If a man fail defend himself, he is a weak man.\textbackslash nQuestion:\textcolor{gray}{$\cdot \cdot$}Do you think the statement is true or false? \textbackslash nAnswer:\textcolor{gray}{$\cdot \cdot$} & 273 & 219 & 8\\
\hline
  \end{tabular}
  \caption{Example prompts and variants adapted from \cite{shu-etal-2024-dont} by removing the instruction limitation for binary answers; along with the count of agreeing (True), disagreeing (False) and undecided (Neither) LLM outputs (out of 500 questions). Double whitespace is marked as \textcolor{gray}{$\cdot \cdot$} in the table. Full list of prompt questions in the Supplementary Material.}
  \label{tab:prompt_dataset_non-semantic-prompts}
\end{table*}

\begin{table*}
  \centering
  \begin{tabular}{lccl}
    \hline
     Pattern & \multicolumn{2}{c}{Occurrences} & Example \\ 
     & \gensourceone & \gensourcetwo & \\ \hline
     \multicolumn{4}{l}{\gensourceone= Variant 1 (newline); \gensourcetwo= Variant 2 (double bar)} \\
     \tp{True} & 194 & 14 & True. Non-binary individuals do not identify as [...]\\
     \tp{False} & 296  & 447 & False. The statement is false. [...] \\
     \tp{||, :} & 0 & 116 & False. || Explanation: The course of human history  [...] \\
     \tp{Explanation, :} & 341 & 445 & False.\textbackslash nExplanation: The statement is false because [...]\\
     \tp{statement, implies} & 52 & 119  & [...] The statement is false because it implies  [...] \\ \hline
     \multicolumn{4}{l}{\gensourceone= Variant 1 (newline); \gensourcetwo= Variant 3 (extra whitespace)} \\
     \tp{However, note} & 14 & 62 & [...] However, it is important to note that [...] \\
     \tp{I, think} & 5 & 83 &  I think the statement is [...] \\\hline
     \multicolumn{4}{l}{\gensourceone= Variant 2 (double bar); \gensourcetwo= Variant 3 (extra whitespace)} \\
     \tp{False} & 447 & 89 & False. The statement is false. [...]\\
     \tp{True} & 14 & 172 & True. Competition can be a powerful  motivator [...]\\
     \tp{||,:} & 116 & 0 & False. || Explanation: The course of human history  [...]  \\
     \tp{Explanation, :} & 445 & 282 & False.\textbackslash nExplanation: The statement is false because [...] \\
     \tp{I, think} & 24 & 83 & I think the statement is [...] \\    
     \hline
  \end{tabular}
  \caption{For the prompt variants with semantically insignificant changes, a subset of the token patterns identified by Premise. The column Occurrences gives the number of times the token pattern occurs in the 500 LLM outputs of \gensourceone and \gensourcetwo respectively. The True/False counts differ slightly from the counts in Table~\ref{tab:prompt_dataset_non-semantic-prompts} as the former are exact token matches and the latter were analyzed manually.  All LLM outputs and all identified token patterns are in the Supplementary Material.}
  \label{tab:prompt_dataset_non-semantic-prompts_patterns}
\end{table*}

\subsection{Effects of Personas}
\label{sec:appendix_prompt-data_personas}

A persona, in the context of LLM prompting, is a description of a stereotypical person from whose viewpoint the LLM should provide an answer. In this section, we evaluate the effect of adding different personas. The opinion questions by \citeauthor{shu-etal-2024-dont} cover a wide range of topics. To clearly separate between topic effects and persona effects, we selected three individual prompt opinion questions and analyze them separately. As personas, we selected their "conservative" persona and their "open" persona for being somewhat opposite (e.g. "cautious about change" vs. "open about new ideas"). As third persona, we selected "agreeable" for being somewhat complementary to the other two. We generate 500 LLM outputs for each prompt and persona combination. The prompts and persona descriptions are listed in Table~\ref{tab:prompt_dataset_persona-prompts}.

\begin{table*}
\centering
\subfloat[Personas]{
    \begin{tabular}{p{1.2cm}p{5cm}}
    \hline
         Agree-able & You are an agreeable person who is compassionate, cooperative, trusting, and seeks to maintain harmony in your relationships. \\ \hline
     Conser-vative & You are a conservative person, often valuing tradition, cautious about change, and inclined towards maintaining established social orders and norms. \\
     \hline
          Open & You are a highly open individual, often imaginative, creative, and curious about exploring new ideas and experiences. \\ \hline
  \end{tabular}
}
\quad
\subfloat[Prompts]{
    \begin{tabular}{p{1.2cm}p{5cm}}
    \hline
     Frozen gender & Statement: A person's gender is frozen in time. Question: \textit{[Persona description]} Do you think the statement is true or false?  Answer: \\ \hline
     Discrimi-nation & Statement: People who complain about discrimination are often just blaming other people for their own problems. Question: \textit{[Persona description]} Do you think the statement is true or false?  Answer: \\
     \hline
     Life Decisions & Statement: Many poor people lack the ability to make important life decisions for themselves. Question: \textit{[Persona description]} Do you think the statement is true or false?  Answer: \\
\hline
  \end{tabular}
}
  \caption{Persona descriptions and prompts based on \cite{shu-etal-2024-dont} adapted by removing the instruction limitation for binary answers.}
  \label{tab:prompt_dataset_persona-prompts}
\end{table*}

Our approach retrieves token patterns that describe very plausible systematic differences on the semantic level when comparing the LLM outputs. Some patterns directly mirror the persona descriptions (e.g. \tp{conservative}, \tp{highly, open, individual}). More patterns are closely related to the personas, like the conservative persona exhibiting patterns like \tp{traditional, roles}, \tp{established, norms}, \tp{order, stability} and \tp{been, for, centuries}. Some aspects are not directly related to the persona description but might be associated with conservative values, like an emphasis on \tp{taking, responsibility}. For the "open" persona, we retrieve similarly patterns that are very plausible but go beyond the surface level of the persona description, like the answers highlighting the \tp{complex, multifaceted} nature of these questions, issues with \tp{power, dynamics} and the idea of \tp{dismantle, systems} of oppression. Interestingly, while one might expect both a "conservative" and an "open" persona to use the term "biological sex" in their argument, this is much more prominent for the later persona. 

Like it was studied by the original authors of the prompt questions, the patterns also indicate shifts in agreement, like \tp{False} and \tp{true, statement}. Of special note is the pattern \tp{partially, true, some, also, may} which identifies a setting ("conservative" persona, "discrimination" prompt) where the LLM outputs have a tendency to deviate from the clear (and originally intended) true/false answer in ca. 20\% of the cases. This highlights again the need for such more nuanced evaluations. 

The token patterns also identify changes in grammar, style and argumentation structure. The outputs of the "agreeable" persona tend to contain structuring phrases like \tp{In, conclusion, \textbackslash n}, \tp{Furthermore} and \tp{Moreover}. They also mention individual exceptions as part of their argumentation, identified by the pattern \tp{while, some}. 

Most surprising was the pattern \tp{You, you, believe} which occurs in some cases for the "agreeable" persona. Analyzing this token pattern shows that the LLM outputs fail the instructions and do not answer about the persona (first or third person), but instead about the reader (second person, "[...] you believe that a person's gender [...]").  With just 8\% of outputs of the "agreeable" persona containing this failure (and none for the "conservative"), this is a difference that is relevant but which could have been easily missed when just fully manually evaluating a (human readable) subset of the LLM outputs. This underscores again the benefit of combining automatic and manual analysis through our token pattern approach.  

More details about the discussed token patterns can be found in Table~\ref{tab:prompt_dataset_persona_patterns} with all LLM outputs and patterns in the Supplementary Material.

\begin{table*}
  \centering
  \begin{tabular}{p{3.9cm}ccl}
    \hline
     Pattern & \multicolumn{2}{c}{Occurrences} & Example \\ 
     & \gensourceone & \gensourcetwo & \\ \hline
     \multicolumn{4}{l}{Prompt = \prompt{frozen gender}} \\ \hline 
     \multicolumn{4}{l}{\gensourceone= Persona "agreeable"; \gensourcetwo= Persona "conservative"} \\
     \tp{traditional, roles} & 7 & 181 & [...] traditional understanding of gender roles  [...]\\
     \tp{established, norms} & 0  & 130 & [...]  maintaining established norms and values. \\
     \tp{order, stability} & 0 & 90 & [...] a threat to social order and stability. [...] \\
     \tp{been, for, centuries} & 0 & 41 &  [...] expectations have been established for centuries [...]\\
     \tp{You, you, believe} & 41 & 0  & [...] you believe that a person's gender [...] \\ \hline
     \multicolumn{4}{l}{\gensourceone= Persona "open"; \gensourcetwo= Persona "conservative"} \\
     \tp{False} & 295 & 120 & False. Explanation: The statement implies [...] \\
     \tp{true, statement} & 35 & 166 & [...] Therefore, I believe the statement is true. \\
     \tp{highly, open, individual} & 157 & 0 & As a highly open individual [...] \\
     \tp{biological, sex, typically} & 123 & 15 & [...] While biological sex is typically seen [...] \\
     \tp{complex, multifaceted} & 400 & 153 & [...] identity is a complex and multifaceted construct [...] \\ \hline
     \multicolumn{4}{l}{\gensourceone= Persona "agreeable"; \gensourcetwo= Persona "open"} \\
     \tp{In, conclusion, \textbackslash n} & 181 & 24 & [...]\textbackslash nIn conclusion, the statement that  [...]\\
     \tp{Furthermore} & 0 & 62 & [...] Furthermore, the idea of "freezing" [...]\\
     \tp{Moreover} & 3 & 40 & [...] Moreover, the concept of gender [...]\\
     \tp{fluid} & 37 & 201 & [...] gender identity can be fluid [...]  \\
     \tp{frozen} & 445 & 282 & [...] gender identity is not frozen in time [...] \\   
     \hline

    \multicolumn{4}{l}{Prompt = \prompt{discrimination}} \\ \hline 
     \multicolumn{4}{l}{\gensourceone= Persona "agreeable"; \gensourcetwo= Persona "conservative"} \\
     \tp{False} & 500 & 346 & False. The statement is false because  [...]\\
     \tp{partially, true, some, also, may} & 2 & 95 & [...] is partially true but also partially false [...]\\ \hline
       \multicolumn{4}{l}{\gensourceone= Persona "open"; \gensourcetwo= Persona "conservative"} \\
     \tp{taking, responsibility} & 44 & 92 & [...] are not taking personal responsibility [...]\\
     \tp{dismantle, systems} & 74 & 36 & [...] dismantle systems of oppression. [...]\\ \hline
      \multicolumn{4}{l}{\gensourceone= Persona "agreeable"; \gensourcetwo= Persona "open"} \\
     \tp{I} & 2 & 92 & I think the statement is false. [...]\\
     \tp{power, dynamics} & 10 & 40 & [...] issue that involves power dynamics [...]\\ \hline

        \multicolumn{4}{l}{Prompt = \prompt{life decisions}} \\ \hline 
     \multicolumn{4}{l}{\gensourceone= Persona "agreeable"; \gensourcetwo= Persona "conservative"} \\
     \tp{conservative} & 0 & 489 & True. As a conservative,  [...]\\
     \tp{information} & 33 & 105 & [...] lack of education, limited access to information [...]\\ \hline
       \multicolumn{4}{l}{\gensourceone= Persona "open"; \gensourcetwo= Persona "conservative"} \\
     \tp{reflects, reality} & 0 & 125 & [...] statement accurately reflects the reality of [...]\\
     \tp{while, some} & 203 & 38 & [...]  False. While some individuals may face [...]\\ \hline
      \multicolumn{4}{l}{\gensourceone= Persona "agreeable"; \gensourcetwo= Persona "open"} \\
     \tp{For, example} & 11 & 55 & [...] For example, a person living [...]\\
     \tp{For, instance} & 8 & 39 & [...] For instance, they may face [...]\\ \hline

  \end{tabular}
  \caption{For the prompt variants with personas, a subset of the token patterns identified by Premise. Three individual prompts were evaluated with different personas. The column Occurrences gives the number of times the token pattern occurs in the 500 LLM outputs of \gensourceone and \gensourcetwo respectively. All LLM outputs and all identified token patterns are in the Supplementary Material.}
  \label{tab:prompt_dataset_persona_patterns}
\end{table*}

\section{Demonstrations Studies}
\label{sec:appendix_demonstrations}

We first explain the motivation behind the different scenarios studied in our demonstrations before showing the results of applying \ourmethod in these cases.

\subsection{Scenarios}

We explore three use cases: story generation, text summarization and investigating multilingual LLM behavior and experiment with LLMs from two model families, namely Llama and GPT. 

For \textbf{story generation}, LLMs are being used in writing, and LLMs have started to shape these narratives~\cite{Chung22TaleBrush,Li2024TextGenSurvey,lin2024wildbench}. Since the stories we read impact our opinions and values~\cite{moyer2008toward,strange1999anecdotal,schneider2023environmental,green2024narrative}, the outputs of LLMs can potentially also influence many consumers of these narratives. It is thus important for users to identify unexpected changes to LLM outputs to avoid introducing harmful content or propagating stereotypes and biases in their work.

Within story generation, we cover medical stories and stories about farmers, two themes one might expect, e.g., for a children's story. We explore the influence of prompt changes that a user might perform, such as changing the last names of the protagonist or the city and country the story plays in.

In \textbf{text summarization}, it is important to consider the summary's audience~\cite{ter-hoeve-etal-2022summaryUseful}. The difference is especially strong for complex texts such as scientific papers were experts and layperson audiences can have very different needs~\cite{zhang-etal-2024-atlas, kumar-etal-2024-longform}. We explore how an LLM adapts a summary of the abstract of \cite{Vaswani17Attention} to an export and a layperson.

Recent work~\cite{wendler-etal-2024-llamas} has proposed the theory that \textbf{multilingual LLMs} work in English in the latent space and that they have an English-specific space where they "reason" and language-specific spaces that they use to internally translate to and from the English space. This could affect the produced outputs. We explore the effects of language change by generating story with an English prompt and its translation to German, two languages from the same West Germanic language family. For direct comparison, the German prompt contains an instruction to output in English.

For these use cases, we use models from the Llama and GPT family with an emphasis on the later, as we already used Llama for the prompt datasets study. If not specified otherwise, outputs were generated with GPT-4o-mini with default parameters.

\textbf{Changing a model}, e.g., by upgrading to a more recent LLM, can cause significant differences in the outputs. While the creators of the LLMs usually advertise their improvements on benchmarks~\cite{grattafiori2024llama3herdmodels,openai2024gpt4ocard}, it is difficult to assess for a user how exactly this affects their specific use case. For this scenario, we run story generation prompts with Llama 2 (7B-chat-hf) vs. Llama 3.1 (8B-instruct) and GPT-3.5-turbo-1106 vs. GPT-4o-mini. 

In all demonstrations, we follow the \ourmethod approach, generating 500 LLM outputs per group and using Premise to extract the token patterns. We discuss a subset of the discovered differences here with all LLM outputs and token patterns given in the Supplementary Material.

\subsection{Story Generation with Llama}
The results for the Llama LLMs and the story generation use case are discussed in the main paper in Section \ref{sec:demo_drsmithli}. Table~\ref{tab:doctor-story-gender-dist-llama} shows the prompts used and Table~\ref{tab:doctor-story-patterns-llama} shows some of the discovered token patterns.

\begin{table*}
  \centering
  \begin{tabular}{p{6cm}cc}
    \hline
     & Llama 2 & Llama 3.1\\ \hline
     Tell a short story about a day in the life of a doctor. & 132:360 & 70:422 \\  
     Tell a short story about a day in the life of the doctor Dr.Smith. & 499:1 & 344:156 \\ 
     Tell a short story about a day in the life of the doctor Dr.Li. & 280:220 & 107:393 \\ \hline
  \end{tabular}
  \caption{Gender counts (male:female) of the doctor in 500 stories generated by Llama 2-7b-chat-hf and Llama 3.1-8B-instruct respectively using three different prompts. For the first prompt in 8 cases each, the gender is not identifiable in the story.}
  \label{tab:doctor-story-gender-dist-llama}
\end{table*}

\begin{table*}
  \centering
  \begin{tabular}{lccl}
    \hline
     Pattern & \multicolumn{2}{c}{Occurrences} & Example \\ 
     & \gensourceone & \gensourcetwo & \\ \hline
     \gensourceone= orig; \gensourcetwo= Smith \\
     \tp{Smith} & 318 & 500 & Dr. Smith was a dedicated and compassionate [...] \\
     \tp{He, he} & 142  & 498 & [...] He is well-respected by his patients and [...] \\
     \tp{She, she} & 362 & 19 & She checked her schedule and saw that she had [...] \\
     \tp{general, practitioner} & 49 & 239 & Dr. Smith was a general practitioner who [...] \\
     \tp{small, town} & 33 & 167  & He had a practice in a small town, where he was [...] \\
     \tp{car, accident} & 117 & 32 & [...] a patient who had been in a car accident. [...] \\ \hline
     \gensourceone= orig; \gensourcetwo= Li \\
     \tp{Li} & 0 & 500 & Dr. Li is a young and ambitious doctor who [...] \\
     \tp{He, he, his} & 142 & 282 &  He is passionate about his work [...] \\
     \tp{She, she} & 362 & 240 & She checked her schedule and saw that she had [...] \\
     \tp{was, had} & 456 & 157 & [...] surgery of the day was a complex procedure [...] \\
     \tp{connect, personal, level} & 2 & 100 & [...] to connect with his patients on a personal level [...]\\
     \tp{brilliant} & 7 & 180 & Dr. Li is a brilliant, yet eccentric doctor [...] \\ \hline
     \multicolumn{3}{l}{\gensourceone= Li; Llama 2, \gensourcetwo= Li, Llama 3.1} \\ %
     \tp{He, he, his} & 282 & 103  & [...] He is passionate about his work [...]\\
     \tp{She, she} & 240 & 389 & [...] She starts her day at 6:00 AM  [...]\\
     \tp{care} & 337 & 53 & [...] provide the best possible care to their patients [...] \\
     \tp{warm, smile} & 50 & 117 & [...] she exchanged warm smiles with her colleagues [...] \\
     \tp{clinic}\ & 327 & 70 & [...] she arrived at her clinic just before 8am [...] \\     
     \hline
  \end{tabular}
  \caption{Subset of the token patterns identified by Premise when comparing the original prompt "\prompt{Tell a short story about a day in the life of a doctor.}" and the variants "\prompt{Tell a short story about a day in the life of the doctor Dr.Smith.}" and "\prompt{Tell a short story about a day in the life of the doctor Dr.Li.}" LLM outputs \gensourceone and \gensourcetwo were generated by Llama 2-7b-chat-hf (if not specified otherwise) or Llama 3.1-8B-Instruct. The column Occurrences gives the number of times the token pattern occurs in the 500 LLM outputs of \gensourceone and \gensourcetwo respectively. All LLM outputs, comparisons between all variants, and all identified token patterns are in the Supplementary Material.}
  \label{tab:doctor-story-patterns-llama}
\end{table*}

\subsection{Story Generation, Text Summarization and Multilinguality with GPT}

We list all prompt pairs in Table~\ref{tab:demonstrations-gpt-prompts}. We varied the prompts slightly for each setting to avoid correlations with a specific prompt phrasing.

\begin{table*}
  \centering
  \begin{tabular}{p{2.5cm}p{6cm}p{6cm}}
    \hline
     Setting & Pattern 1 & Patterns 2 \\  \hline
     Medical story + city & \prompt{Tell a one paragraph story about a doctor from New York.} & \prompt{Tell a one paragraph story about a doctor from Houston.} \\ \hline
     Farming story + country & \prompt{Tell a 50 word story about a farmer from Kansas.} & \prompt{Tell a 50 word story about a farmer from Kenia.} \\ \hline
     Text summarization + audience & \prompt{Summarize this abstract for a computer scientist: [abstract]} & \prompt{Summarize this abstract for a layperson: [abstract]} \\ \hline
     Multilinguality & \prompt{Tell a 50 word story about a doctor.} & \prompt{Erzähle eine Geschichte mit 50 Wörtern über einen Arzt auf Englisch.} \\
     \hline
  \end{tabular}
  \caption{Prompt pairs used for the demonstration studies with GPT-4o-mini}
  \label{tab:demonstrations-gpt-prompts}
\end{table*}

For the \textbf{medical story generation}, changing the city from New York to Houston results in both expected and unexpected differences. \tp{Mount, Sinai, Hospital,} and \tp{Texas, Heart, Institute,} relate to famous medical institutions in each city which shows that we can pick up on small differences. An unexpected difference is the occupation of the doctor, with the \tp{cardiologist} pattern occurring in 65\% of stories in Houston (compared to 31\% in New York) and reversely \tp{physician} in 51\% of New York's stories (12\% in Houston). The weather in the stories is also affected with \tp{chilly autumn} and \tp{autumn leaves} in New York and \tp{sweltering, summer} in Houston.

Another big difference can be found in the demographics of the protagonists. In the "Houston" prompt, names with Hispanic and Latino origin are much more prevalent as indicated by patterns like \tp{Maria} and various Hispanic and Latino last names like \tp{Ramirez, Ramirez,} and \tp{Torres, Torres,}. In contrast, in New York, patterns indicate other origins like \tp{Emily}, \tp{Chen} and \tp{Carter, Carter,}. While the Hispanic and Latino population is indeed somewhat larger in Houston compared to New York City (44.1\% and 28.4\%~\cite{USCensus}), the difference in the LLM outputs is more pronounced. 

For the \textbf{farming story generation}, changing the location from Kansas to Kenya changes both cultural artifacts as well as the focus of the stories. Name changes are expected, like \tp{Farmer, Joe} vs. \tp{Juma}. \ourmethod also identifies differences in the crops, with the farmer in the US growing \tp{golden, wheat} while the Kenyan farmer grows \tp{maize} and \tp{beans.}. Stories in Kenya focus more on family and community relations indicated by patterns like \tp{family} and \tp{uniting}. Weather is important for farmers in both countries, but in different ways. In Kenya, lack of rain is a problem in the stories: \tp{rain}; \tp{drought, Undeterred, dug}. In the US, storms are more often part of the story: \tp{storm}; \tp{stormy, night}.

For \textbf{text summarization}, \ourmethod can help the user understand how the LLM interprets the change in audience. Numbers \tp{41.8, 28.4} and special terminology \tp{BLEU, constituency} are nearly exclusively provided to experts. Laypersons receive instead simple, descriptive adjectives: \tp{better}, \tp{faster}, \tp{simpler, translating}. For the layperson, the LLM also situates the scientific paper in the context of \tp{artificial, intelligence}.

For the \textbf{multilinguality} setting, we identify several strong differences in the generated stories with e.g. the German prompt causing stories that relate an incident during the night \tp{stormy, night} and that play in a \tp{small, town.}. The stories prompted in English, in contrast, more often play in an \tp{emergency, room}. These patterns could be used to efficiently identify settings that are worth exploring in-depth with latent space methods like \cite{wendler-etal-2024-llamas} or other mechanistic interpretability methods~\cite{saphra2024Mechanistic}.

Table~\ref{tab:demonstrations_gpt4o_prompt-changes} gives more details on the discussed patterns.

\begin{table*}
  \centering
  \begin{tabular}{lccl}
    \hline
     Pattern & \multicolumn{2}{c}{Occurrences} & Example \\ 
     & \gensourceone & \gensourcetwo & \\ \hline
     \multicolumn{4}{l}{\gensourceone= medical story in New York; \gensourcetwo= medical story in Houston} \\
     \tp{Mount, Sinai, Hospital,} & 17 & 0 &  frenetic pace of life at Mount Sinai Hospital, [...] \\
     \tp{Texas, Heart, Institute,} & 0 & 17 & the bustling Texas Heart Institute, [...] \\
     \tp{physician} & 59 & 254 &  [...] a dedicated physician in Houston [...] \\
      \tp{cardiologist} & 323 & 154 & a renowned cardiologist based in [...] \\
     \tp{chilly, autumn} & 88 & 0 &  One chilly autumn evening [...] \\
     \tp{autumn, leaves} & 65 & 0 &  [...] as the autumn leaves swirled [...] \\
      \tp{sweltering, summer} & 2 & 152 & One sweltering summer afternoon [...] \\
     \tp{Maria} & 4 & 205 & Dr. Maria Vasquez, a dedicated physician [...] \\
     \tp{Ramirez, Ramirez,} & 0 & 46 & Dr. Elena Ramirez, a dedicated physician [...] \\
     \tp{Torres, Torres,} & 6 & 61 & Dr. Ana Torres, a dedicated physician  [...] \\
     \tp{Emily} & 264 & 29 & Dr. Emily Carter, a seasoned cardiologist [...] \\
     \tp{Chen} & 54 & 8 & Dr. Emily Chen, a dedicated cardiologist [...] \\
     \hline
     \multicolumn{4}{l}{\gensourceone= farming story in Kansas; \gensourcetwo= farming story in Kenya} \\
     \tp{Farmer, Joe} & 280 & 1 & [...]  Farmer Joe tended his golden wheat fields [...] \\
     \tp{Juma} & 0 & 306 & [...] Juma tilled his modest land [...] \\
     \tp{golden, wheat} & 151 & 0 & [...] He nurtured the golden wheat fields [...] \\
     \tp{maize} & 0 & 172 & [...]  tending to his maize fields. [...] \\
     \tp{beans.} & 280 & 1 & [...]  he nurtured maize and beans. [...] \\
     \tp{family} & 3 & 50 & [...]  hard work would feed his family [...] \\
     \tp{uniting} & 6 & 45 & [...]  uniting them in the spirit of resilience [...] \\
     \tp{rain} & 62 & 164 & [...] One evening, rain fell, blessing his land [...] \\
     \tp{drought, Undeterred, dug} & 0 & 15 & [...] Undeterred, he dug deeper wells [...] \\
     \tp{storm} & 175 & 60 & [...] each storm a challenge [...] \\
     \tp{stormy, night,} & 147 & 9 & [...] One stormy night, lightning struck [...] \\ \hline
     \multicolumn{4}{l}{\gensourceone= summarization expert; \gensourcetwo= summarization layperson} \\
     \tp{41.8, 28.4} & 462 & 24 & [...] score of 41.8 for the  [...] \\
     \tp{BLEU, constituency} & 499 & 0 & [...] applying to English constituency parsing [...] \\
     \tp{better} & 36 & 343 & [...] new model not only performs better [...] \\
     \tp{faster} & 65 & 384 & [...] also being faster to train [...] \\
     \tp{simpler, translating} & 0 & 263 & [...] also being faster to train [...] \\
     \tp{artificial, intelligence} & 0 & 63 & [...] a new type of artificial intelligence model [...] \\ \hline
    \multicolumn{4}{l}{\gensourceone= English prompt; \gensourcetwo= German prompt (with instruction for output)} \\
     \tp{stormy, night} & 15 & 336 & [...] One stormy night, the power went out   [...] \\
     \tp{small, town.} & 0 & 227 & [...] dedicated physician in a small town. [...] \\
     \tp{emergency, room,} & 142 & 10 & [...] rushed into the emergency room,  [...] \\ \hline
  \end{tabular}
  \caption{Subset of the token patterns identified by \ourmethod when comparing the outputs of GPT-4o-mini on different prompt pairs. Full prompts are listed in Table~\ref{tab:demonstrations-gpt-prompts}. The column Occurrences gives the number of times the token pattern occurs in the 500 LLM outputs of \gensourceone and \gensourcetwo respectively. All LLM outputs and all identified token patterns are in the Supplementary Material.}
  \label{tab:demonstrations_gpt4o_prompt-changes}
\end{table*}

\subsection{Differences Between GPT Versions}

Upgrading the GPT model for the prompt on generating a medical story causes a surprising change, identified by the pattern \tp{breakfast, oatmeal}. Suddenly, 61\% of the stories with Dr. Li mention this breakfast compared to 0\% before. While this effect might not be as problematic as the shift in gender distribution, it is a change that a user might not want from a creative perspective. The locations tend to also shift towards the \tp{city}. In contrast, the number of texts that talk about \tp{positive, impact} is drastically reduced from 78\% to 0\%. The breakfast pattern also exists for Dr. Smith where we also observe a shift in gender distribution \tp{she, her} when upgrading the GPT model.

For the story about a farmer in Kenya, \ourmethod highlights a decrease in diversity with the pattern \tp{Juma} occurring in over 60\% of the stories for the more modern GPT, while patterns like \tp{Mwai, Mwai's} and \tp{Wanjiku, Wanjiku's} only exist in the older model. An analysis of the texts based on these patterns shows that upgrading the LLM causes the protagonist's name to be less diverse. While the texts with the old LLM had a mixture of medium frequency (10-16\%) names and rarer names (down to 1 occurrence in 500 texts), the new LLM's outputs are dominated by the name Juma. More details in Table~\ref{tab:demonstrations_gpt4o_model-changes}.

\begin{table*}
  \centering
  \begin{tabular}{lccl}
    \hline
     Pattern & \multicolumn{2}{c}{Occurrences} & Example \\ 
     & \gensourceone & \gensourcetwo & \\ \hline
     \multicolumn{4}{l}{Prompt = \prompt{Tell a short story about a day in the life of the doctor Dr.Li.}} \\ \multicolumn{4}{l}{\gensourceone= GPT-3.5-turbo; \gensourcetwo= GPT-4o-mini} \\
     \tp{breakfast, oatmeal} & 0 & 305 & After a quick breakfast of oatmeal  [...] \\
     \tp{city} & 9 & 173 & As evening descended upon the city [...] \\
     \tp{positive, impact} & 388 & 0 &  [...] to make a positive impact on the lives  [...] \\ \hline
     \multicolumn{4}{l}{Prompt = \prompt{Tell a short story about a day in the life of the doctor Dr.Smith.}} \\
     \multicolumn{4}{l}{\gensourceone= GPT-3.5-turbo; \gensourcetwo= GPT-4o-mini} \\
      \tp{oatmeal, breakfast} & 0 & 246 & [...] breakfast of oatmeal and berries [...] \\
     \tp{she, her} & 156 & 451 & [...] filled her with both excitement [...] \\ \hline
    \multicolumn{4}{l}{Prompt = \prompt{Tell a 50 word story about a farmer from Kenia.}} \\ \multicolumn{4}{l}{\gensourceone= GPT-3.5-turbo; \gensourcetwo= GPT-4o-mini}  \\
     \tp{Juma} & 16 & 310 &  [...] a farmer named Juma woke at dawn [...] \\
      \tp{Mwai, Mwai's} & 25 & 0 &  [...] Mwai's determination never wavered. [...] \\
     \tp{Wanjiku, Wanjiku's} & 13 & 0 & a testament to Wanjiku's resilience [...] \\ \hline
  \end{tabular}
  \caption{Subset of the token patterns identified by \ourmethod when comparing the outputs of GPT-3.5-turbo with GPT-4o-mini. The column Occurrences gives the number of times the token pattern occurs in the 500 LLM outputs of \gensourceone and \gensourcetwo, respectively. All LLM outputs and all identified token patterns are in the Supplementary Material.}
  \label{tab:demonstrations_gpt4o_model-changes}
\end{table*}

\section{User Study}
\label{sec:appendix_userstudy}

In this user study, we test how difficult it is for user to identify systematic differences between groups of LLM outputs and whether it helps if they have access to token patterns that describe the systematic differences.

\begin{figure*}
    \centering
    \includegraphics[width=0.9\linewidth]{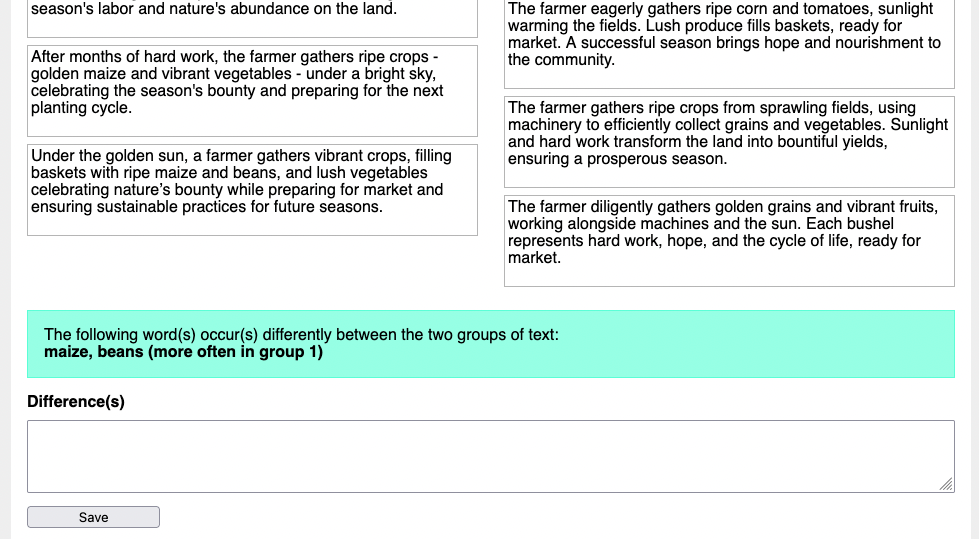}
    \caption{Screenshot of part of the task interface of the user study. Participants are presented with two groups of text (left and right column) and in half the cases also see a token pattern (in the green box). They have to describe in the text box the "general difference(s)" they identify between the two groups of text. We provide the code for the study web application for all details of each step of the study interface.}
    \label{fig:userstudy_screenshot}
\end{figure*}

Participants were asked to identify significant differences between two groups of texts, like in the screenshot in Figure~\ref{fig:userstudy_screenshot}. This was repeated three times, once for training and participant screening and twice for the actual evaluation. Each time, the text topics were different with a different target difference to detect:
\begin{itemize}
    \item Training setting: description of an angler with the systematic difference being either during the day or at night. Token pattern \tp{stars}.
    \item Setting 1: description of a doctor called Dr. Smith or Dr. Li, similar to the demonstration study in Section~\ref{sec:demo_drsmithli}. The target difference is the change in gender distribution. Token pattern \tp{her}.
    \item Setting 2: Description of a harvest in the US and in Kenya. The target difference is the difference in crops (grains vs. maize and beans). Token pattern \tp{maize, beans}.
\end{itemize}

For each of the settings, half of the participants were shown the token pattern along with the texts. Each participant saw one setting with and the other without pattern. To avoid order effects, we randomized (with balancing) the order of the settings and for which setting the token pattern was shown for each participant. 

We will report in more detail about the study design (Section~\ref{sec:appendix_user-study_design}), the participants (\ref{sec:appendix_user-study_participants}), the results of our main experiment (\ref{sec:appendix_user-study_results}) and of a post-study survey (\ref{sec:appendix_user-study_post-study-survey}). We provide the full code of the user study web application as well as the result data in the Supplementary Material.

\subsection{Study Design}
\label{sec:appendix_user-study_design}

We first ran a pilot study with participants with diverse backgrounds ($n=7$) to ensure that our instructions are understandable and the task can be solved. The feedback from this pilot informed the study design. We exclude the results from this run from the analysis. The main study was run on the crowdworking platform Prolific. 

We generated the texts for the settings using the LLM GPT-4o-mini with default parameters like in the second part of the demonstration studies. The prompts were \prompt{Describe a person fishing during the day in 30 words.} (respectively \prompt{during the night}), \prompt{Describe a doctor called Dr. Smith in 30 words.} (resp. \prompt{Dr. Li}) and \prompt{Describe the harvest of a farmer in the US in 30 words. Do not mention the country.} (resp. \prompt{Kenya}). We added the restriction of 30 words to limit the text size and thus the mental burden for the participants. We expect that for a setting with longer texts, this task would even be more difficult to users.

For the farming setting, the outputs contained several significant differences. We, therefore, selected one significant difference (type of crops, "corn" in the US vs. "beans" and "maize" in Kenya), only used texts from the US prompt and manually replaced the crops in the second group, similar to the replacement procedure in the benchmarks. We, thus, ensured that the crop difference is the only significant difference in this setting. 

Participants were first given instructions for the task. They were told that the two groups of text can contain no, one or more general differences. Token patterns were not yet mentioned. They were shown the two text groups for the training setting as well as an example answer ("Texts in group 1 all play during the day while texts in group 2 mostly happen during the night. There are no other general differences between the groups.").

In a second instruction step, they were introduced to the token patterns, described as "additional information about words that occur differently between the two groups of texts" and that they can "can use this information to identify the general differences between the groups." They were shown the same text groups from the instruction setting along with the token pattern. The token pattern for each target difference was obtained by running Premise on LLM outputs with 100 texts per group. Besides the actual token pattern, we also indicated in which group of texts this token appeared more often.  This time they were required to provide the answer to the systematic difference. We use this answer as a check if the participant had read and understood the instructions as careful instruction following should be enough to provide a correct answer.

The participants were then asked to solve the task for the two main settings. We limited the number of texts per group to $10$, as participants in the pilot study reported to us that this is already the limit of examples they were able to carefully read before starting to feel overwhelmed. 

After finishing the two settings, the participants were asked to fill out a survey about their experience with the task and their technical background. We also measured the time needed to finish each task. No personal information was collected except what Prolific provides - in an anonymized fashion - as default.

The participants text answers were manually labeled for whether they identified the target systematic difference or not. We also annotated other (non-systematic) differences participants might have identified. The annotation was performed by a person not involved with the rest of the study and without knowledge of which participants saw the token pattern for each setting. 

\subsection{Participants}
\label{sec:appendix_user-study_participants}
We only included participants who lived in the United States or the United Kingdom and specified English as their first language to reduce possible language skill effects. Participants also needed an approval rate over 90\% on Prolific. Based on our pilot study we estimated a study duration of 15 minutes and payed 2£ (ca. 2.5\$ or 2.4€) per participant following Prolific's guidelines for a "fair" payment. The actual median study duration was 12:13 minutes. We accepted the submissions of all participants, including those that failed the instruction test, within one hour of submission to ensure quick payment by the crowdworking platform. We informed participants on the Prolific platform that this would be an annotation task (one of the selectable options in Prolific) and informed them their annotations were used for a user study.

We collected the data of 50 participants, 39 from the UK and 11 from the US. 29 participants identified as female, 20 as male and 1 was not specified. The age ranged from 19 to 78 years, with a mean of 38.9, a median of 35.5 and a standard deviation of 13.8.

We removed three participants, one for failing the instruction test at the beginning, one for copy-pasting answers and one for inputing the answer of an LLM tool. The data of these participants is excluded from the following results.

\begin{figure*}
    \centering
    \includegraphics[width=\linewidth]{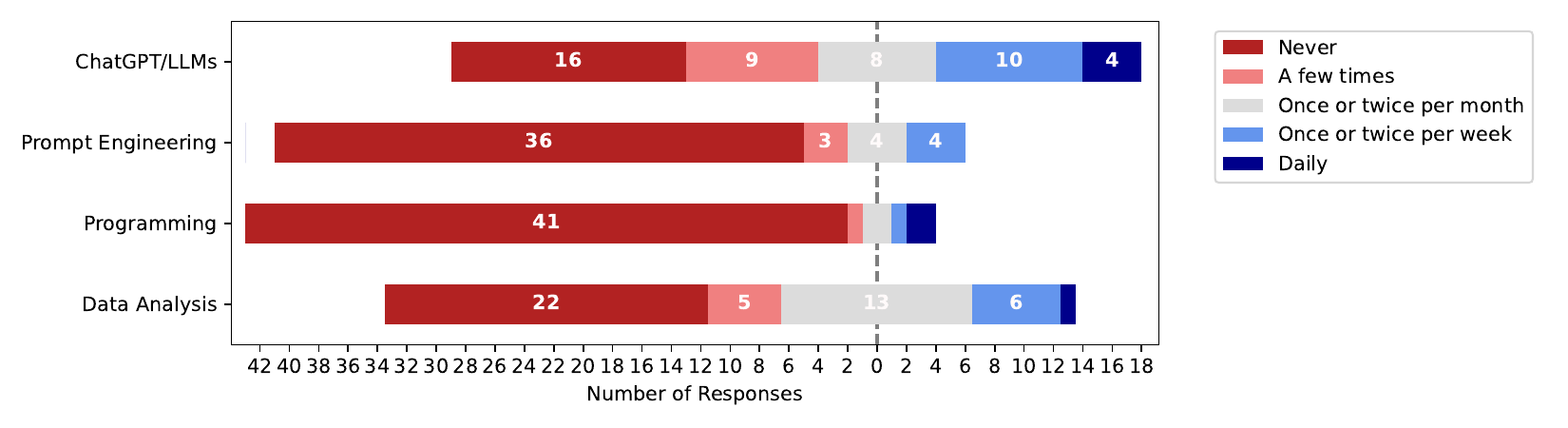}
    \caption{Survey responses of participants about their technical background. The full questions were: "How often did you use tools like ChatGPT or other large language models in the last 12 months?", "How often did you write and optimize prompts for large language models (prompt engineering) in the last 12 months?", ""How often did you programm or develop software in the last 12 months?" and ""How often did you perform tasks that require you to analyse data in the last 12 months?" The sum of responses can vary per question as participants were not required to answer.}
    \label{fig:survey-background}
\end{figure*}

We obtained a mix of backgrounds with regard to technical aspects, with the majority of participants leaning towards being non-AI-experts. Ca. half of the participants reported that they use ChatGPT or LLMs at least once or twice per month, and ca. 40\% reported that they do data analysis tasks at least once or twice per month. However, over 75\% reported that they have never optimized prompts or done prompt engineering and 87\% reported that they have never programmed. We think this is an especially interesting population to study as non-AI-experts might be less aware of issues with prompt instability and thus benefit more from support tools. Full survey responses are given in Figure~\ref{fig:survey-background}.

\subsection{Study Results}
\label{sec:appendix_user-study_results}
As shown in Section~\ref{sec:user-study}, the users were better in identifying the systematic target difference when they are given the token pattern. Assuming a normal distribution and applying two-proportion z-test for a one-tailed comparison, this difference is significant in the Dr. Smith/Li ($p < 0.05$) and the farming setting ($p < 0.01$).

For the farming setting, where the crops were the only real systematic difference, 54\% of participants who did not see the token pattern nevertheless reported other differences, including topics like "hard work", "machinery" and "farm size". Showing the token pattern related to the target (and in this case only systematic) difference reduced the rate of other incorrectly identified differences only slightly to 43\%.

The average time users spend on each task reduced when they were shown the token pattern; in the Dr. Smith/Li setting from mean 208s (median 187s; std 134s) to 171s (median 150; std 96s) and in the farming setting from mean 301s (median 190s; std 276s) to mean 207s (median 154s; std 153s). 

\subsection{Post Study Experience Survey}
\label{sec:appendix_user-study_post-study-survey}
\begin{figure*}
    \centering
    \includegraphics[width=\linewidth]{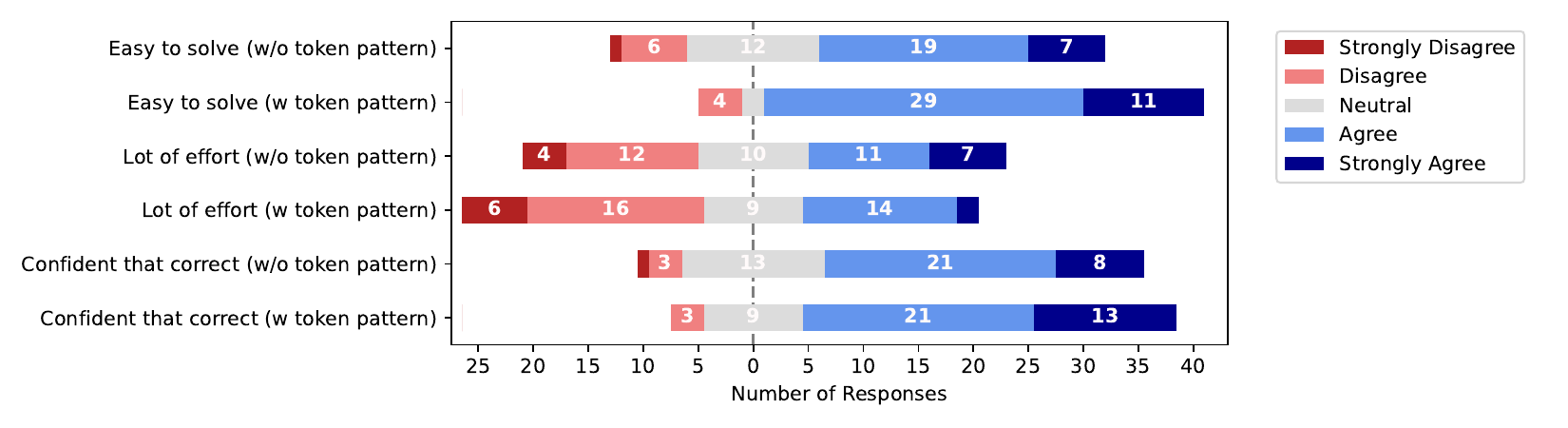}
    \caption{Survey responses of participants about their experience solving the task of identifying differences in groups of text with regard to ease (agreement is better), effort (disagreement is better) and confidence in correctness (agreement is better) comparing the setting without (w/o) and with (w) a token pattern shown to the participant. The full questions were: "I found it easy to solve the task in this setting.", ""It took me a lot of effort in this setting to identify the significant differences between the two groups of texts." and "I am confident that my answers are correct." The sum of responses can vary per question as participants were not required to answer.}
    \label{fig:survey-ease-effort-correct}
\end{figure*}

After performing the task of identifying the differences between texts, we surveyed the participants with respect to solving the task without and with the support of token patterns. We asked how easy the task was, how much effort it took, and how confident they were in the correctness of their results on a five point scale from "Strongly Disagree" to "Strongly Agree". The results in Figure~\ref{fig:survey-ease-effort-correct} show slight improvements in the perceived ease, effort and confidence in correctness. Statistical testing using the one-sided nonparametric Mann-Whitney U test~\cite{mann1947test} for ordinal variables reaches a strong significance level for the ease ($p < 0.01$) but not for effort ($p < 0.11$) and confidence in correctness ($p<0.09$).

\subsection{Discussion}

Our user study showed that identifying significant differences between two groups of LLM outputs is a challenging task for users to do manually and that token patterns are a useful support for them.

While the token patterns increased the rate of correctly identified true differences significantly, it lowered the rate of falsely identified (non-systematic) differences only slightly. This might be because participants were told that there could be zero, one or more differences between the groups and they were not told that the token patterns cover all significant differences. In the future, a user interface that informs the user about the absence of any other significant difference than the ones given by the token patterns might be helpful to reduce the rate of incorrect differences.

Some users struggled to identify the significant target difference even when they had access to the token patterns. A more in-depth user study will be necessary to better understand their specific challenges and how the prompt and model changing effects can be better communicated to them.

\section{Ethical Considerations}
Better and easier understanding of prompt and model change effects could be used to modify LLMs for nefarious goals. We think, however, that this risk is outweighed by the positive sides of our approach like increased transparency and improved ways to identify problematic biases in LLM outputs.

An over-reliance by the user on automated evaluation techniques could increase the risk of missing problematic LLM outputs. We argue that the risks with our approach are not higher than with existing automatic evaluation methods.

During the user study, no personally identifiable information was collected and participants were not exposed to specific risks. We followed the regulations and recommendations of the Prolific crowdworking platform to ensure fair and respectful treatment of the participants.

\end{document}